\pdfoutput=1
\documentclass[11pt]{article}

\usepackage[]{acl}

\usepackage{times}
\usepackage{latexsym}

\usepackage[T1]{fontenc}
\usepackage[utf8]{inputenc}

\usepackage{microtype}

\usepackage{inconsolata}

\usepackage{xspace}
\usepackage{tabularx}
\usepackage{algorithm}
\usepackage{algorithmic}
\usepackage{graphicx}
\usepackage{booktabs}
\usepackage{amssymb}
\usepackage{pifont}
\usepackage{caption}
\usepackage{comment}
\usepackage{array}
\usepackage{longtable}
\usepackage{multirow}
\usepackage{listings}
\usepackage{amsmath}
\usepackage{xcolor}
\usepackage{geometry}
\usepackage{enumitem}
\usepackage{booktabs}
\usepackage{arydshln}

\definecolor{ForestGreen}{RGB}{34,139,34}
\definecolor{BrickRed}{rgb}{.72,0,0}
\definecolor{LakeBlue}{RGB}{0,61,153}
\usepackage{bbding}
\usepackage{soul}
\definecolor{myYellow}{rgb}{0.9,0.9,1}
\sethlcolor{myYellow}
\usepackage{wrapfig}

\title{Boosting Language Models Reasoning with Chain-of-Knowledge Prompting}

\author{Jianing Wang\textsuperscript{$\diamondsuit$}\thanks{~~J. Wang and Q. Sun contributed equally to this work.}, Qiushi Sun\textsuperscript{$\heartsuit$}\footnotemark[1], Xiang Li\textsuperscript{$\diamondsuit$}\thanks{~~Corresponding author.}, 
Ming Gao\textsuperscript{$\diamondsuit$} \\
\textsuperscript{$\diamondsuit$}East China Normal University
\textsuperscript{$\heartsuit$}The University of Hong Kong \\
\texttt{lygwjn@gmail.com},
\texttt{qiushisun@u.nus.edu}, \\
\texttt{\{xiangli, mgao\}@dase.ecnu.edu.cn}
} 

\begin{document}
\maketitle
\begin{abstract}
Recently, Chain-of-Thought (CoT) prompting has delivered success on complex reasoning tasks, which aims at designing a simple prompt like ``\emph{Let's think step by step}'' or multiple in-context exemplars with well-designed rationales to elicit Large Language Models (LLMs) to generate intermediate reasoning steps.
% However, the reasonableness and accurateness of generated reasoning chains still remain elusive: the rationale is a textual hint and often comes with mistakes that make it hard to verify.
However, the generated rationales often come with hallucinations, making unfactual and unfaithful reasoning chains.
To mitigate this brittleness, we propose a novel Chain-of-Knowledge (CoK) prompting, where we aim at eliciting LLMs to generate explicit pieces of knowledge evidence in the form of structure triple.
This is inspired by our human behaviors, i.e., we can draw a mind map or knowledge map as the reasoning evidence in the brain before answering a complex question.
% , and then verify it by checking the evidence path with the off-the-shelf knowledge base.
Benefiting from CoK, we additionally introduce a F\textsuperscript{2}-Verification method to estimate the reliability of the reasoning chains in terms of \emph{factuality} and \emph{faithfulness}.
For the unreliable response, the wrong evidence can be indicated to prompt the LLM to rethink.
Extensive experiments demonstrate that our method can further improve the performance of commonsense, factual, symbolic, and arithmetic reasoning tasks~\footnote{The code and data are available at~\url{https://github.com/wjn1996/Chain-of-Knowledge}}.

% \url{https://anonymous.4open.science/r/Chain-of-Knowledge-36EE}.}.
% ~\footnote{The code and data will be released upon acceptance.}.
\end{abstract}

\section{Introduction}
\label{sec:introduction}
\begin{figure*}
    \centering
	\includegraphics[width=\linewidth]{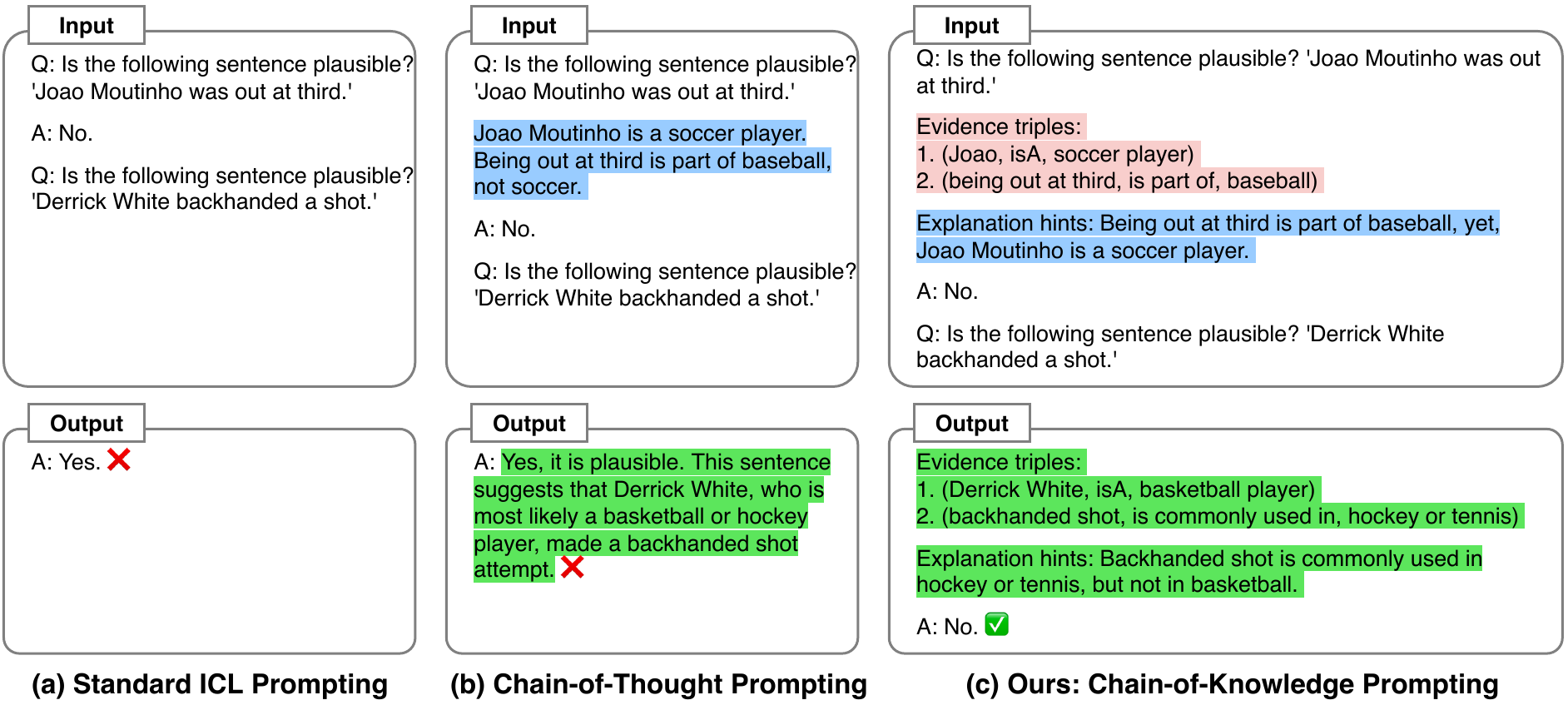}
	\caption{Comparison of three prompting methods: (a) ICL, (b) Chain-of-Thought (CoT), and (c) Chain-of-Knowledge (CoK) solving a StrategyQA question.}
% \label{fig:overview}
 \label{fig:overview}
 \vspace{-1em}
\end{figure*}

Large Language Models (LLMs) have succeeded in advancing the state-of-the-arts for many Natural Language Processing (NLP) tasks
~\citep[][\emph{inter alia}]{Brown2020Language, Jack2021Scaling, Romal2022LaMDA, Aakanksha2022PaLM, scao2022bloom, zhang2022opt, bai2022training, Touvron2023LLaMA},
benefiting from the ultra-large-scale training corpora and computation resources.
To unleash the LLMs' power of adaptation on unseen tasks without any parameter updates, in-context learning (ICL) has become one of the flourishing research topics, aiming at generating the prediction by conditioning on a few labeled exemplars (Figure~\ref{fig:overview} (a))~\cite{Shin2022On, Zhao2021Calibrate, Liu2022What, Lu2022Fantastically, Dong2023A}.

A series of recent works have explored that LLMs can spontaneously decompose the complex multi-step problem into intermediate reasoning chains
% ~\footnote{The reasoning chains consist of rationales and the final answer.}
, elicited by a simple prompt like ``\emph{Let's think step by step}'' or well-designed demonstrations with human-annotated rationales, which are called chain-of-thought (CoT) prompting (Figure~\ref{fig:overview} (b))~\cite{Wei2022Chain, Kojima2022Large, Wang2022Self, Zhou2022Least, yao2023react}. This finding is intriguing and has been sensational because CoT may mainly specify an output space/format that regularizes the model generation to look step-by-step while being in order and relevant to the query~\cite{Wang2022Towards, Min2022Rethinking}.

% Despite these incredible capabilities on complex reasoning NLP tasks
% Despite incredible performances, we find that there are still two challenges for existing CoT methods:
% 1) the reasoning chain is only a textual rationale, it is hard to verify its reasonableness and accurateness~\footnote{By intuition, textual reasoning chains are often vague and elusive.}, and

% sqs: incredible -> impressive; make unfactual or unfaithful reasoning chains -> susceptible to generating xxx

Despite impressive performances, 
current LLMs are susceptible to generating hallucination~\cite{ji2023hallu, Zhang2023Siren}, 
along with providing unfactual or unfaithful reasoning chains that inevitably lead to a wrong conclusion~\citep{wang2023factuality}.
Take Figure~\ref{fig:overview} as an example. Given a query ``\emph{Is the following sentence plausible `Derrick White backhanded a shot.'}'' from StrategyQA~\cite{geva2021strategyqa}, the standard ICL and CoT make a wrong answer.
One of the reasoning steps ``Derrick White is most likely a hockey player'' is fake (In fact, Derrick White is a basketball player), making the unfactual inference towards the question.
In addition, 
the response may be unfaithful when the LLM generates logically sound reasoning chains while still providing an incorrect answer.

% sqs: generates correct reasoning chains but outputs a wrong answer. -> generates logically sound reasoning chains while still providing an incorrect answer.

% most -> biggest
% sqs: is nonnegligible for the final answer -> has a non-trivial impact on the final answer

% To eliminate these concerns, we propose a novel \textbf{Chain-of-Knowledge (CoK)} prompting to boost the LLM's reasoning capability by combining structure knowledge (concerned by symbolicism, e.g., knowledge graph) and neural decoding model (concerned by connections, e.g., GPT-3).

% sqs: eliminate -> address
To address these concerns, we propose a novel \textbf{Chain-of-Knowledge (CoK)} prompting method to boost the LLM's reasoning capability by a series of exemplars that combine explicit structure knowledge evidence with textual explanations.
To elaborate, 
CoK prompting consists of two compositions (Figure~\ref{fig:overview} (c)), i.e., evidence triples (CoK-ET) and explanation hints (CoK-EH), where CoK-ET is a list of structure triples can reflect the overall reasoning evidence from the query towards the answer and CoK-EH is the explanation of this evidence.
% ~\footnote{Explanation hint is similar to textual rationale in CoT.}.
% 此处需要说明，CoK的构建中，正确与否不会影响模型决策
To construct in-context exemplars with the CoK prompt, we first sample $K$ labeled examples and each of them can be concatenated with a simple hint ``\emph{Let's think step by step}'' to prompt the LLM to generate reasoning chains. 
Then, we retrieve some structure triples from the external knowledge base (KB) and judiciously manually annotate evidence triples to obtain a well-designed CoK prompt.
% a F$^2$-verification -> an F$^2$-verification
Like standard ICL and CoT, the CoK prompt can be perceived as a rule that regularizes the output space/format and urges LLMs to generate explicit evidence instead of only attempting to generate vague textual reasoning chains. 
Furthermore,
we also propose an F\textsuperscript{2}-Verification strategy to estimate the reliability of the reasoning chains in terms of \emph{factuality} and \emph{faithfulness}, where \emph{factuality} is the quantification of the matching degree between reasoning evidence and ground-truth knowledge, and \emph{faithfulness} is the consistency degree between reasoning evidence and the textual explanation with the final answer.
Particularly for the unreliable response, 
the wrong pieces of evidence can be indicated to prompt the LLM to rethink the problem.
We design a \emph{rethinking algorithm} to reach this goal.

% sqs: (commonsense, factual reasoning) -> (e.g., commonsense & factual reasoning)
% sqs: on some -> across various ...
We have conducted empirical evaluations across various reasoning tasks (e.g., commonsense, factual, arithmetic, and symbolic), showing that CoK prompting with F\textsuperscript{2}-Verification can significantly outperform standard ICL and CoT prompting.
We also integrate CoK prompting with some prevailing strategies, such as self-consistency.
% and graph-of-thought (GoT), 
The results indicate that such CoK can serve as a plug-and-play module to further improve reasoning ability.

% We also explore that our framework exhibits exceptional performances on some non-knowledge-intensive tasks (e.g., arithmetic, symbolic reasoning).

% sqs: Different prompting: standard ICL, chain-of-thought, and our chain-of-knowledge. -> Comparison of 3 prompting methods: (a) standard ICL, (b) Chain-of-Thought (CoT, with manually designed demonstrations), and (c) Chain-of-Knowledge (CoK) solving a strategyQA question.

\section{Related Work}
\label{sec:related_work}
\paragraph{Prompting for LLMs with in-context learning.}
% A series of works focus on tunable prompt-tuning~\cite{Gao2021Making, Raffel2020Exploring, Liu2021GPT, Gu2022PPT} to bridge the gap between pre-training and fine-tuning. 
In-Context Learning (ICL) is the task of causal language modeling, allowing LLMs to perform zero/few-shot learning with a well-designed text-based prompt~\cite{Brown2020Language, Aakanksha2022PaLM, Touvron2023LLaMA, Romal2022LaMDA, Dong2023A}. 
ICL can bypass the model parameter update and achieve the salient performance by conditioning on a few labeled examples.
Previous works have explored some impact facets of ICL. For example, the input-output mapping and the template format~\cite{pan2023what, Min2022Rethinking, Yoo2022Ground}, 
% sqs: add new icl paper
the different selection and permutations of the exemplars~\cite{Lu2022Fantastically}.
To improve ICL's effectiveness, some novel methods have been proposed, involving meta-learning~\cite{Chen2022Meta, Min2022MetaICL}, prompt and exemplars engineering~\cite{Liu2022What, Liu2023Pre}. 
% obtaining LM preferences by the interaction between inputs and in-context examples~\cite{ye2023ceil},
% finding appropriate example organization through self-adaption mechanism~\cite{wu2023self},
% and prediction calibrating~\cite{Zhao2021Calibrate}, etc.
% Different from them, our work focuses on the reasoning ability of LLMs in two aspects, including reasoning prompt design and output verification.

% sqs: add ye2023ceil(icml23) + wu2023self(acl23)

% sqs: instead of them 感觉怪怪的，感觉In our study, we concentrate on the reasoning capabilities...会好些？@wjn

% \begin{figure*}[th!]
\begin{figure*}
    \centering
	\includegraphics[width=\linewidth]{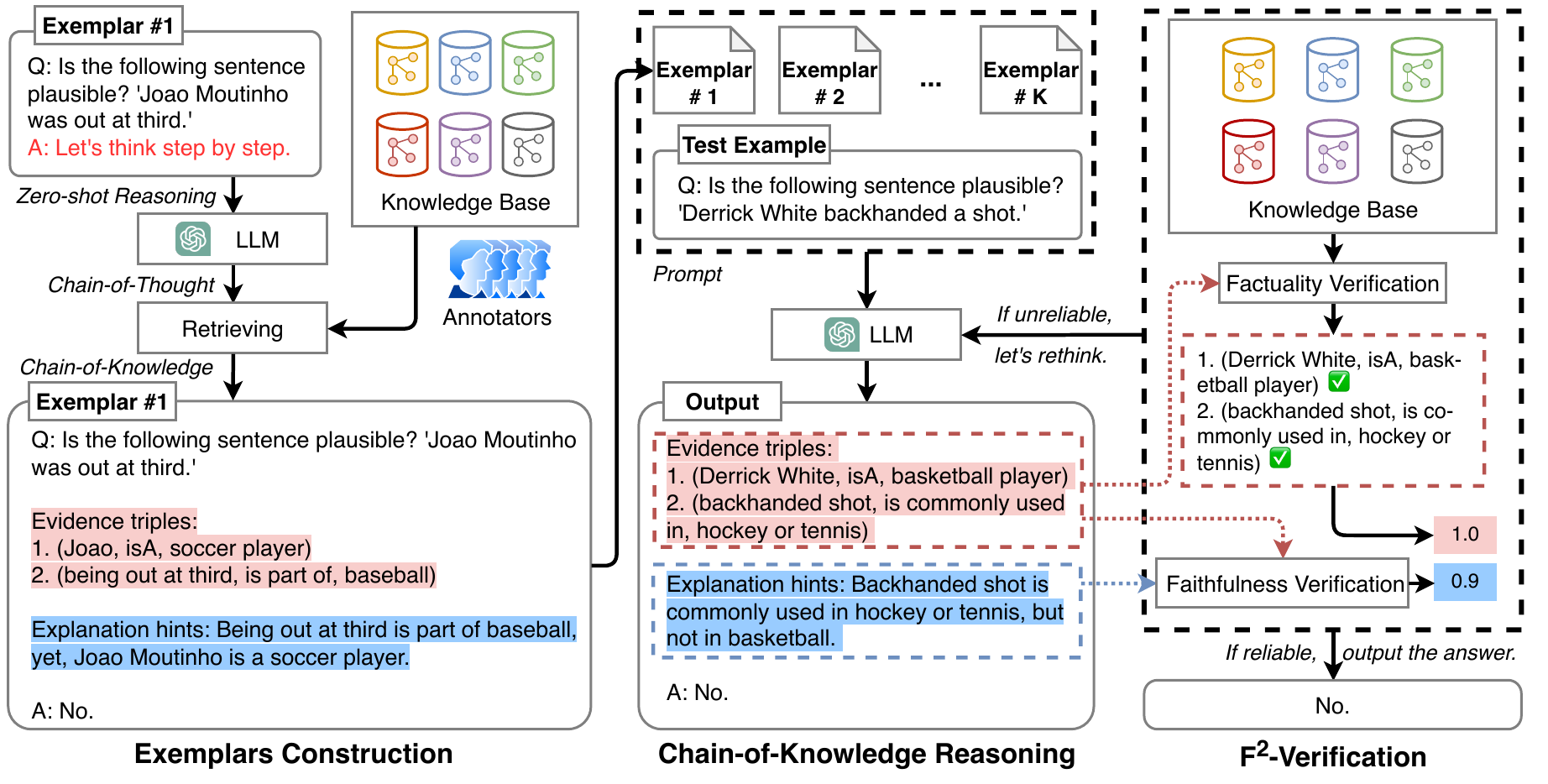}
	\caption{The proposed framework. We first construct exemplars with chain-of-knowledge (CoK) prompts. Then, the CoK prompts can be used to let the LLM generate reasoning chains, including evidence triples, explanation hints, and the final answer. Lastly, we estimate the reliability of reasoning chains in terms of \emph{factuality} and \emph{faithfulness}, and the unreliable ones will be rethought.}
 \label{fig:framework}
 \vspace{-1.em}
\end{figure*}

\paragraph{Chain-of-thought prompting elicits reasoning.}
Recently, CoT prompting has been presented to leverage reasoning and interpretable information to guide LLMs to generate reliable and explainable responses \cite{Wei2022Chain}.
A series of CoT-enhanced methods are proposed to further improve the reasoning ability~\cite{Kojima2022Large, Huang2022Large, Wang2022Self, Si2022Prompting, Wang2022Rationale, Zhou2022Least, Zhang2022Automatic, fu2023complexcot, Besta2023Graph}. 
For example,
\citet{Wang2022Self} introduce \emph{Self-consistency} to suppress the wrong rationales problem by marginalizing out the sampled reasoning paths to find the most consistent answer.
% \emph{Least-to-most}~\cite{Zhou2022Least} is proposed to tackle complex reasoning tasks by decomposing the original query into some easy-to-hard subproblems, making rationales more effective.
% \emph{Program of Thoughts}~\citet{chen2022program} {employ} a program interpreter to delegate the computation, aiming at decoupling computational processes from reasoning and understanding.
\citet{fu2023complexcot} and \citet{Besta2023Graph} proposed logical thinking graph to let LLMs better reasoning.
% \citet{shum2023automatic} circumvent manual involvement by augmenting rational chains from small datasets.
% In addition, a series of works improve the reasoning performance by the verification, which is similar to ours. For example,
\citet{Lyu2023Faithful} translates the complex problem into interleaved natural language or programming language to make the reasoning chains faithful. 
\citet{Li2022On} and \citet{Yao2023Tree} introduce coarse or fine-grained labels to verify the reasoning chains.
Differently, we focus on alleviating the hallucination in terms of the factuality and faithfulness of the reasoning chains.

\section{Methodology}
\label{sec:methodology}
% Preliminary

% Chain-of-Thought prompting aims to decompose the complex problem into multiple intermediate steps and solve it before reaching the final answer~\cite{Wei2022Chain}. 
% This simulates human beings’ own thought processes when meeting complex problems, such as multi-step reasoning tasks or math word problems. 
% Figure~\ref{fig:overview} (medium) demonstrates an example of producing an answer with a CoK prompt. 
% CoT是以纯文本形式存在的，从而引导大模型生成文本式的推理。但是这种推理通常并不清晰，从而很难被验证是否正确
The generated reasoning chains elicited by CoT prompting sometimes come with mistakes, ultimately leading to hallucinated answers.
We attribute this challenge to the textual reasoning chain:
% which is sometimes vague and hard to verify. 
% In other words, if one question is hard, 
% \emph{LLMs may forcibly generate a textual rationale that conforms to the prompt format of CoT but is logically ambiguous or even wrong to achieve the purpose of prevarication and try to deceive the questioner}.
\emph{LLMs may forcibly generate a textual rationale that conforms to the prompt format of CoT but is logically ambiguous and reaches the wrong answer}.
To address this challenge,
% 因此，我们从两个方面来改进CoT
% ============= add in final-vesion ===
% To address this challenge, we thus consider two core aspects. 
% 1) Elicitation format of the prompt. Text-only reasoning chains are not enough to unleash LLMs to generate reliable and concrete reasoning processes. Inspired by the triple structure in the KB, we need to enhance the prompt with structured features. 
% 2) Post-verification. LLMs are usually not capable of inspecting what answers they had responded to, indicating us to leverage external knowledge to make verification.
% Through the consideration of these facets, 
% ============= add in final-vesion ===
% Based on these considerations, 
we provide our specific solution on how to boost LLM's reasoning ability in two corners: the reasoning elicitation format of the prompt and the knowledge-enhanced post-verification.
The overview of the framework is shown in Figure~\ref{fig:framework}.
% \begin{itemize}
%     % \setlength{\itemsep}{0pt}
%     % \setlength{\topsep}{1pt}
%     \item Better elicitation format of the demonstations. Inspired by the triple structure in the knowledge graph, we introduce chain-of-knowledge (CoK) prompting to decompose the complicated problems into multiple explicit evidence triples besides textual hints.
%     \item Post-verification. We further propose a F$^2$-Verification strategy to estimate the reliability of the reasoning chain, outputting final answers before verification.
% \end{itemize}

\subsection{Chain-of-Knowledge Prompting}
\label{sec:chainofknowledge_prompting}
% wjn：如果能够通过理论说明引入结构的prompt更好，就好了。探索理论方面可以在appendix部分添加。
% We first explore a better elicitation prompt format. 
 % \sqs{\st{As we all know,} 
It is widely recognized that reasoning can be modeled as induction and deduction on the existing knowledge system~\cite{goswami2002inductive}. This is inspired by human behaviors that draw a mind map or knowledge map to analyze the question and find the correct path to the answer.
Fortunately, we can 
% \sqs{\st{borrow the idea} 
adopt the concept of the triple in the KB, which can be viewed as a ``(subject, relation, object)'', to formalize the explicit evidence of the reasoning chains. To elaborate, we propose \textbf{Chain-of-Knowledge} (CoK) prompting to facilitate a better elicitation prompt for LLMs, which consists of two main ingredients, i.e., evidence triples (CoK-ET) and explanation hints (CoK-EH).
CoK-ET represents a list of multiple triples and each of them represents the knowledge evidence probed from LLMs to support the step-by-step thinking process.
CoK-EH denotes the explanation of the reasoning chain, which is similar to CoT. 
% \sqs{\st{chain-of-thought prompt} vanilla CoT}.
% For example, in Figure~\ref{fig:overview}, we can urge the LLM to generate two explicit shreds of evidence ``(Derrick White, isA, basketball player), (basketball shot, is commonly used in, hockey or tennis)'' and a textual explanation to support the final answer ``No''.
Take Figure~\ref{fig:overview} as an example, we can urge the LLM to generate explicit shreds of evidence to support the final answer.

\subsection{Exemplars Construction}
\label{sec:exemplars_construction}
% sqs: As just discussed in previous works -> Building upon the insights of previous studies

Building upon the insights of previous studies~\cite{Zhang2022Automatic, Min2022Rethinking, Wang2022Self}, 
the performance of ICL hinges on the annotated rationale.
This indicates that the key challenge of CoK prompting lies in constructing textual rationales with their structure evidence triples.
% 注意：有工作表明，demonstrate的正确与否对具体样本的生成影响不大
As shown in Figure~\ref{fig:framework}, we first perform exemplars construction to obtain a well-designed task-specific prompt. 
Specifically, we follow \cite{Wei2022Chain, Wang2022Self} to randomly select $K$ questions as the basic demonstrations.
To automatically obtain CoK-EH, we follow \cite{Kojima2022Large} to generate a textual rationale for each question via zero-shot CoT with a simple prompt ``\emph{Let's think step by step}''. 
Another challenge is how to obtain annotated CoK-ET that better expresses the textual rationale. 
To figure it out, we first follow~\cite{Pan2022Knowledge} to construct a KB $\mathcal{K}$ from six domains, involving \textit{dictionary}, \textit{commonsense}, \textit{entity}, \textit{event}, \textit{script}, and \textit{causality}, which are in the form of triple.
% We then aim at retrieving some relevant knowledge triples from this KB. 
We then directly use the retrieving tool proposed by~\cite{Pan2022Knowledge} to retrieve some candidate triples.
% During retrieval, given a textual explanation (i.e., CoK-EH), we encode it with the RoBERTa~\cite{Liu2019RoBERTa} and then retrieve the most relevant knowledge triples using the maximum inner product search tool SCaNN~\cite{Guo2020Accelerating}.
Finally, we invite 5 people as professional annotators to manually design the corresponding CoK-ET based on the retrieved triples~\footnote{In fact, during the exemplars construction, the generated textual reasoning chains and retrieved triples could have some mistakes.
% , inevitably leading to the wrong guideline for model prediction.
Fortunately, we found that there is no strong connection between the reasoning validity of both CoK-ET and CoK-EH and the performance of the model predictions, which is similar to findings in~\cite{Wang2022Towards}.
We will bring detailed discussions at Section~\ref{sec:discussions}.}.
% We provide two strategies to figure it out. On the one hand, we invited 5 teachers or students as professional annotators to manually designed the corresponding evidence triples. Thus, each exemplar has five different evidence triples, representing the knowledge from different human minds.
% On the other hand, we follow~\cite{Pan2022Knowledge} to construct a KB $\mathcal{K}$ from six domains, involving \textit{dictionary}, \textit{commonsense}, \textit{entity}, \textit{event}, \textit{script}, and \textit{causality}, which are in the form of triple. Then, we design one or more knowledge-specific strategies to generate key-value pairs from the original knowledge pieces, and encode all keys into dense vectors using the best sentence encoder MPNet~\cite{Song2020MPNet}. 
% During retrieval, given a textual reasoning chain (e.g., CoK-EH), we encode it with the same sentence encoder model and then retrieve the most relevant knowledge triple using the maximum inner product search tool SCaNN~\cite{Guo2020Accelerating}. We use five different random seeds to obtain different evidence triples lists.
More details can be found in Appendix~\ref{app:cok_construction}.

% \subsection{F$^2$-Verification}

% sqs: ↓这种方式字体和正文能保持一致，学长看要不要用@wjn
\subsection{F\textsuperscript{2}-Verification} 
\label{sec:f2ver}

After the exemplars construction, we can obtain $K$ annotated data $\mathcal{E}=\{(Q_i, T_i, H_i, A_i)\}_{i=1}^{K}$.
Symbolically, $Q_i$, $H_i$ and $A_i$ represent the input query, the explanation hint, and the final answer of the $i$-th exemplar, respectively; each of them is the token sequence.
$T_i$ denotes the list of evidence triples, which contains multiple knowledge triples, i.e., $T_i=\{(s_{ij}, r_{ij}, o_{ij})\}_{j}$, where $s_{ij}$, $r_{ij}$ and $o_{ij}$ are subject, relation and object, respectively.
Given a test query input $\hat{Q}_i$, we can directly choose one permutation of $\mathcal{E}$ and concatenate them with this test query into a linear sequence $\hat{I}_i=[\mathcal{E}; \hat{Q}_i]$ to prompt the LLM to generate the prediction, i.e., $\hat{y}_{ik}=\arg\max_{\sigma(\hat{y}_{i(\leq k)})} P(y|\hat{y}_{i(<k)}, \mathcal{E}, \hat{Q}_i)$, where $P(y|\cdot)$ is the prediction distribution, $\hat{y}_{ik}$ is the $k$-th token, $\sigma(\cdot)$ denotes the decoding strategy (e.g., temperature sampling, beam search, and nucleus sampling), $[\cdot;\cdot]$ is the concatenation operation.

Due to the well-designed format of the demonstrations, the final prediction derived from the LLM $\hat{y}_{i}$ consists of a list of evidence triples $\hat{T}_i$, a sequence of explanation hints $\hat{H}_i$ and the final answer $\hat{A}_i$.
However, LLMs may generate hallucinated rationales, so the final answer can not be guaranteed. 
We attribute this problem to two factors: 
1) some steps in the rationale may not correspond to the fact, contributing to the wrongness, and
2) the relation between the final answer and the reasoning chains is still ambiguous, making the response unfaithful. 
To alleviate these drawbacks, we propose F\textsuperscript{2}-Verification to estimate the answer reliability towards both \textbf{F}actuality and \textbf{F}aithfulness~\footnote{We find that~\cite{he2022rethinking} also proposed rethinking and retrieving processes to reduce wrongness, different from them, we focus on fine-grained detection and injection with the proposed CoK prompt.}.

\paragraph{Factuality Verification.} We first verify the factuality, which can be viewed as the matching degree between each generated evidence triple and the ground-truth knowledge from KBs~\footnote{We assume that the knowledge from the KB is correct and up-to-date.}.
% Benefiting from the format of the chain-of-knowledge prompt, the reasoning chains can be decomposed into multiple explicit evidence triples that have the same structure as the KB.
% Therefore, we can verify each evidence triple supervised by the ground-truth knowledge in the external KB~\footnote{We assume that the knowledge from the KB is correct and up-to-date.}.
% sqs: updated -> up-to-date?
% , so, the factuality of each query $\hat{Q}_i$ can be simply quantified as the matching degree of $\hat{H}_i$ supervised by the KB.
Concretely, we first define a function $f_{v}(\hat{r}_{ij}| \hat{s}_{ij}, \hat{o}_{ij}, \mathcal{K})$ to represent the factuality of each evidence.
We design two different strategies of $f_v$. 
1) Exact verification. We can retrieve all relevant triples based on the subject $\hat{s}_{ij}$ and object $\hat{o}_{ij}$, and then find whether the generated relation $\hat{r}_{ij}$.
i.e., $f_{v}(\hat{r}_{ij}| \hat{s}_{ij}, \hat{o}_{ij}, \mathcal{K})=\mathbb{I}((\hat{s}_{ij}, \hat{r}_{ij}, \hat{o}_{ij})\in\mathcal{K})$ exists.
2) Implicit verification. For a triple that does not exist in KB, it could be corrected.
Thus, we can transform the factuality verification into a graph completion task that predicts whether the triple is true.
% sqs: factuality verifying -> verification? and faithfulness verifying
For simplicity, we use TransR~\cite{Lin2015Learning} to pre-train the KB $\mathcal{K}$ and use the off-the-shelf energy function to assign a score for each evidence triple, i.e., $f_{v}(\hat{r}_{ij}| \hat{s}_{ij}, \hat{o}_{ij}, \mathcal{K})=||\mathbf{s}_{ij}^{(r,c)}+\mathbf{r}^c-\mathbf{o}_{ij}^{(r,c)}||_2^2+\alpha ||\mathbf{r}^c-\mathbf{r}_{ij}||_2^2$, where $\alpha>0$ is the balancing factor, $||\cdot||_2^2$ is Frobenius norm.  
$\mathbf{s}_{ij}^{(r,c)}=\mathbf{s}_{ij}\mathbf{M}_r$ and $\mathbf{o}_{ij}^{(r,c)}=\mathbf{o}_{ij}\mathbf{M}_r$  denote the projection representations of the subject $s_{ij}$ and object $o_{ij}$ in the relation space $r$, respectively. 
$\mathbf{s}_{ij}, \mathbf{r}_{ij}$, and $\mathbf{o}_{ij}\in\mathbb{R}^{d}$ are the $d$-dimension embeddings of $\hat{s}_{ij}$, $\hat{r}_{ij}$ and $\hat{o}_{ij}$, respectively. 
$\mathbf{r}^c\in\mathbb{R}^{d}$ is the prototype embeddings of the relation $r$. $\mathbf{M}_{r}\in\mathbb{R}^{d\times d}$ is the trainable projection matrix of relation $r$.
We join the two strategies in our framework. If the evidence triple exists in $\mathcal{K}$, we will use an exact verification strategy to assign a score, or we use an implicit verification strategy.
% sqs: exactly -> an exact ...

\paragraph{Faithfulness Verification.} As defined by~\citet{Jacovi2020Towards} and \citet{Lyu2023Faithful}, \emph{if the reasoning process derived from the model can accurately be expressed by an explanation, we call it faithful}. 
Previous works based on chain-of-thought prompting struggle to verify faithfulness due to the lack of sufficient evidence to understand the relation between the explanation and the answer~\cite{ye2022the}.
% sqs: add unreliability explanation cites
So, we propose a faithfulness verification method to find out these cases.
Specifically, given one test query $\hat{Q}_{i}$, a list of evidence triples $\hat{T}_{i}$ and the final answer $\hat{A}_{i}$, we directly concatenate them as a new sequence $\hat{H}'_i$. We leverage the pre-built sentence encoder SimCSE~\cite{Gao2021SimCSE} to calculate the similarity between $\hat{H}'_{i}$ and $\hat{H}_{i}$. 
We denote this function as $f_u(\hat{H}_{i}|\hat{H}'_{i}=[\hat{Q}_{i}; \hat{T}_{i}; \hat{A}_{i}])=SimCSE(\hat{H}_{i}, \hat{H}'_{i})$. 

Finally, for each query $\hat{Q}_i$ we can obtain a score $\mathcal{C}_i$ ($0<\mathcal{C}_i<1$) that represents whether the rationale is reliable towards the answer:
\begin{equation}
\begin{aligned}
\mathcal{C}_{i} &= \gamma\frac{1}{|\hat{T}_{i}|}\sum_{j=1}^{|\hat{T}_{i}|}f_v(\hat{s}_{ij}|\hat{r}_{ij}, \hat{o}_{ij}, \mathcal{K}) \\
&+ (1-\gamma)f_u(\hat{H}_{i}|\hat{H}'_{i}=[\hat{Q}_{i}; \hat{T}_{i}; \hat{A}_{i}]),
\label{eqn:reliability}
\end{aligned}
\end{equation}
where $0<\gamma<1$ is the balancing factor and set to be $0.5$ as default, $|\hat{T}_i|$ is the number of triples.

% @sqs 需要添加 ↓
% 就是之前有很多向大模型注入外部知识的，他们都是在大模型推理之前通过讲知识注入在prompt上来预测的。我们这个认为他们检索到的知识并不一定是大模型需要的
% 因此，我们通过cok来显式让模型生成evidence triple，然后找到那些存在错误的triple，并将对应正确的知识，在rethinking阶段注入进去，从而实现精准注入
% wjn: 这个点是也是一个比较关键的点，但是我可能在全文没有写出这个，你看看结合全文在哪里强调一下
% sqs: 意思大概是↓，我找个合适的地方融合到文章里，或者跟在rethinking后面

\subsection{Rethinking Process}

% \li{Algorithm -> remove}

\label{sec:rethinking_algorithm}
% We design a rethinking algorithm to further improve the performance. 

F\textsuperscript{2}-Verification facilitates us to ensure the factuality and faithfulness of the triples and explanations generated by the model. 
% Moreover,
Beyond the scope of verification,
we can further boost the performance of LLMs by employing a \emph{rethinking process},
as shown in Algorithm~\ref{alg:rethinking}.
% The algorithm pseudo-code Algorithm~\ref{alg:rethinking} can be found in Appendix~\ref{app:rethinking}.
We initialize a reliability threshold $\theta$ ($0<\theta<1$), iteration number $N$, and an unreliability set $U$. 
All the queries in the testing set $\mathcal{D}_{test}$ are initially deemed unreliable.
For each query $\hat{Q}_{i}\in U$ in the $n$-th iteration, we obtain CoK prompt $\hat{I}_i^{(n)}$ by combining the demonstrations $\mathcal{E}$ and the query $\hat{Q}_{i}$.
The prompt can be used to elicit the LLM to generate a list of evidence triples $\hat{T}_i^{(n)}$, 
explanation hints $\hat{H}_i^{(n)}$, and the final answer $\hat{A}_i^{(n)}$.

The proposed rethinking algorithm will allow the LLMs to assess the reliability of the rationales (i.e., $\hat{T}_i^{(n)}$ and $\hat{H}_i^{(n)}$) via calculating the score in Eq.~\ref{eqn:reliability}.
% and then followed by the calibration of those samples deemed unreliable.
% Specifically, 
% for each $\hat{Q}_{i}$,
% the corresponding reliability score $\mathcal{C}_{i}$ is calculated using Equation~\ref{eqn:reliability} on the basis of $\hat{H}'_{i} = [\hat{Q}_{i}, \hat{T}_{i}, \hat{A}_{i}]$ and $\mathcal{K}$.
An entry is no longer considered unreliable if $\mathcal{C}_{i}^{(n)}$ is not below the threshold $\theta$, which subsequently leads to the final answer $\hat{A}_{i}$.
Conversely, 
suppose $\mathcal{C}_{i}^{(n)}$ fails to reach $\theta$. 
In that case, we can select the evidence triples with lower scores and inject the corresponding correct knowledge triples from the KB into the CoK prompt $\hat{I}_i^{(n+1)}$ in the next iteration (Line 12)~\footnote{This is similar to correcting wrong reasoning paths, we ensure that the label is not leaked to the model.}.
% the process is repeated with the regeneration of new reasoning chains.
This dynamic generate-evaluate procedure continues until all entries in $U$ are considered reliable or the maximum number of iterations $N$ is reached. For cases where the maximum number of iterations is reached without any triples' reliability score surpassing $\theta$, triples with the highest reliability scores will be selected for inference (Line 15-17).
% \li{score -> scores}

\begin{algorithm}\small
\caption{Rethinking Process}
\label{alg:rethinking}
\begin{algorithmic}[1]
% \footnotesize
\REQUIRE Exemplars $\mathcal{E}$, testing query set $\mathcal{D}_{test}\leftarrow\{\hat{Q}_i\}_{i=1}^{M}$, KB $\mathcal{K}$, iterator number $N(\geq1)$, reliability threshold $0<\theta<1$.
\STATE Initialize an unreliability set $U\leftarrow\mathcal{D}_{test}$.
\FOR{each iteration $n \leftarrow 1, \cdots, N$}

\FOR{each query $\hat{Q}_{i}$ in $U$}
\STATE Obtain a CoK prompt $\hat{I}_i^{(n)}$. If $n$ is $1$, $\hat{I}_i^{(n)}\leftarrow[\mathcal{E}; \hat{Q}_i]$.
\STATE Generate evidence triple $\hat{T}_{i}^{(n)}$, explanation hint $\hat{H}_{i}^{(n)}$ and answer $\hat{A}_{i}^{(n)}$ from the LLM.
\STATE Calculate reliability score $\mathcal{C}_{i}^{(n)}$ in Eq.~\ref{eqn:reliability}.
\IF{$\mathcal{C}_{i}^{(n)}\geq\theta$}
    \STATE Obtain final answer $\hat{A}_{i}\leftarrow\hat{A}_{i}^{(n)}$.
    \STATE Remove $\hat{Q}_{i}$ from $U$.
    \STATE \textbf{continue}
\ENDIF
\STATE For the evidence triples that $f_v(\hat{r}_{ij}^{(n)}|\hat{s}_{ij}^{(n)}, \hat{o}_{ij}^{(n)}, \mathcal{K})<\theta$, inject the corresponding correct knowledge triples $\hat{T}'_i$ into the prompt, i.e., $\hat{I}_i^{(n+1)}\leftarrow[\hat{I}_i^{(n)}; \hat{T}'_i]$.
\ENDFOR
\ENDFOR
\FOR{each query $\hat{Q}_i$ in $U$}
\STATE Obtain the final answer $\hat{A}_i\leftarrow\arg\max_{\hat{A}_i^{(n)}}\mathcal{C}_{i}^{(n)}$.
\ENDFOR
\STATE \textbf{return} all the answers $\{\hat{A}_{i}\}_{i=1}^{M}$.
\end{algorithmic}
\end{algorithm}

\begin{table*}[t]
\centering
% \vspace{-3mm}
\setlength{\tabcolsep}{1pt}
\vspace{2.8mm}
\resizebox{\textwidth}{!}{
\begin{small}
\begin{tabular}
{lccccccccccccc}
\toprule
\multirow{3}{*}{Model} & \multicolumn{6}{c}{\textit{Commonsense \& Factual}} & \multicolumn{2}{c}{\textit{Symbolic}} & \multicolumn{4}{c}{\textit{Arithmetic}} \\
\cmidrule(r){2-7}
\cmidrule(r){8-9}
\cmidrule(r){10-13}
&Common   &Strategy & OpenBook &ARC-c &Sports &BoolQ &\space Letter\space &\space Coin\space & GSM8K & SVAMP & AQuA & MultiArith  \\
&Sense QA   &QA &QA & & & & & & & & & \\
\midrule
Fine-tuning  & 91.2 & 73.9  & 91.0 & 75.0 & - & 92.4 & - & - & 55.0 & 57.4  & 37.9 & - \\
% \midrule
% \multicolumn{13}{l}{\textit{text-davinci-002 reasoning results}}\\
% \midrule
\midrule
\multicolumn{13}{c}{\textit{text-davinci-002 reasoning results}}\\
\midrule
Zero-Shot SP  &  68.8 & 12.7  & 44.7 & 46.8 & 38.1 & 50.2 & 0.2 & 12.8 & 10.4 & 58.8 & 22.4 & 17.7 \\
Zero-Shot CoT  & 64.6  & 54.8 & 68.4 & 64.7 & 77.5 & 52.7 & 57.6 & 91.4 & 40.7 & 62.1 & 33.5 & 78.7 \\
Few-Shot SP & \textbf{79.5}  & 65.9 & \textbf{76.6} & 68.2 & 69.6 & 53.6 & 0.0 & 49.1 & 15.6 & 65.7 & 24.8 & 33.8 \\
Manual CoT   & 73.5 & 65.4 & 73.0 & 69.9 & 82.4 & 55.0 & 59.0 & 74.5 & 46.9 & 68.9 & 35.8 & 91.7 \\
Auto-CoT & 74.4   & 65.4   & -   & -  & -  & -  &  59.7 &  \textbf{99.9} & 47.9 & 69.5 & 36.5 & 92.0 \\
\hdashline
CoK &  75.4  & 66.6 &  73.9 & 71.1  & 83.2 & 56.8 & 59.4 & 97.4 & \textbf{51.2} & \textbf{69.9} & \textbf{37.8} & \textbf{94.6} \\
CoK + F\textsuperscript{2}-V &  77.3  &  \textbf{67.9}  &  74.8 & \textbf{73.0}  & \textbf{84.1}  & \textbf{59.9}  & \textbf{61.1}  & - & - & - & - & - \\
\midrule
\multicolumn{13}{c}{\textit{gpt-3.5-turbo reasoning results}}\\
\midrule
Manual CoT  & 76.5  & 62.6  & 82.6 & 84.9 & 84.0 & 65.1 & 73.0 & 97.4 & 79.1 & 79.5 & 55.1 & 97.3 \\
Manual CoT + SC  & 78.2  &  63.7 & 85.0 & 86.5 & 86.5 & 66.6 & \textbf{74.5} & 99.0 & 87.6 & 85.0 & 66.8 & 98.8 \\
ComplexCoT  & 75.4  & 62.2  & - & - & - & - & - & - & 79.3 & 77.7 & 56.5 & 95.4 \\
ComplexCoT + SC  & 76.0 & 63.0  & - & - & - & - & - & - & \textbf{89.2} & 85.6 & 65.0 & 98.23 \\
\hdashline
CoK  & 77.1  & 63.8 & 83.5 & 85.7 & 85.9 & 67.9 & 63.1 & 98.0 & 83.2 & 81.4 & 60.2 & 99.0 \\
CoK + SC  & 78.9  &  65.0 & 86.1 & \textbf{87.5}  & 87.4 & 69.4 & 68.3 & \textbf{99.2} & 88.2 & \textbf{86.0} & \textbf{69.7} & \textbf{99.3} \\
CoK + F\textsuperscript{2}-V  & 77.8  & 64.5  & 85.0 & 86.6 & 87.0 & 69.2 & 65.4 & - & - & - & - & - \\
CoK + SC + F\textsuperscript{2}-V  & \textbf{79.3}  & \textbf{66.6}  & \textbf{87.0} & 87.4 & \textbf{87.9} & \textbf{69.9} & 69.7 & - & - & - & - & - \\
\bottomrule
\end{tabular}
\end{small}
}
\caption{Accuracy of~\emph{text-davinci-002} and ~\emph{gpt-3.5-turbo} model over commonsense, factual, symbolic, and arithmetic reasoning tasks.}
\label{tab:main_results}
\end{table*}

\section{Experiments}
\label{sec:experiments}
% In this section, we first describe the experiment settings about tasks, implementation details, and baselines, and then provide the main results.

\subsection{Experimental Setups}
\label{sec:experimental_setups}
% \begin{wrapfigure}{r}{0.5\textwidth}
% % \vspace{-4mm}
% \includegraphics[width=\linewidth]{images/chain-of-knowledge}
% \caption{Example of secret-sharing tensors, revealing tensors, and private addition in \crypten{}.\sqs{test float}}
% \label{fig:example1}
% % \vspace{-4mm}
% \end{wrapfigure}

\paragraph{Tasks and Datasets.}
During experiments, we choose five different kinds of tasks to evaluate the performance of our method. The datasets and corresponding implementation details are shown in the following.
% sqs：似乎可以使用\begin{enumerate}[topsep=1pt,itemsep=0ex]%缩点版面，如果需要的话
% sqs: add datasets, benchmarks and descriptions, subject to change
1) \textbf{Commonsense \& factual reasoning}.
We select CommonsenseQA (CSQA)~\cite{talmor2019commonsenseqa}, StrategyQA~\cite{geva2021strategyqa}, OpenBookQA~\cite{mihaylov2018openbookqa} the AI2 Reasoning Challenge (ARC-c)~\cite{clark2018think}, sports understanding from the BIG-Bench benchmark ~\cite{srivastava2022imitation}, and BoolQ~\cite{clark2019boolq} for evaluating CoK on commonsense and factual reasoning.
% 2) \textbf{Factual reasoning.} For these tasks, both closed-book question answering: HotpotQA~\cite{yang2018hotpotqa}, BoolQ~\cite{clark2019boolq} and knowledge base question answering: WebQuestions~\cite{berant2013semantic} are employed.
2) \textbf{Symbolic reasoning.} Two symbolic reasoning tasks are evaluated in our experiments, specifically, Last Letter Concatenation and Coin Flip tasks~\cite{Wei2022Chain}.
3) \textbf{Arithmetic reasoning.} We use grade school math problems GSM8K~\cite{cobbe2021training}, a challenging dataset over math word problems SVAMP~\cite{patel2021nlp}, and two others AQuA~\cite{Ling2017Program}, MultiArith~\cite{Roy2015Solving} for math problem solving tasks.

\paragraph{Implementation Details.}

% sqs: add robyn2017conceptnet, mostafazadeh2020glucose, wang2021kepler, sap2019atomic
% \vspace{-10pt}
For the LLM, we employ the publicly accessible GPT-3~\cite{Brown2020Language} models, namely, \emph{gpt-3.5-turbo} and \emph{text-davinci-002} with 175B parameters unless otherwise stated.
% We directly follow \cite{Wei2022Chain, Zhang2022Automatic, Wang2022Self, Kojima2022Large} to choose the same labeled examples as the demonstrations.
We use greedy decoding with temperature 0 and max output length 512, keeping consistent with baselines for a fair comparison.
For the datasets from commonsense reasoning and factual reasoning, 
% and natural language understanding, 
the KBs we choose are a combination of Wiktionary~\footnote{\url{https://en.wiktionary.org/wiki/Wiktionary}.}, ConceptNet~\cite{robyn2017conceptnet}, Wikidata5M~\cite{wang2021kepler}, ATOMIC~\cite{sap2019atomic}, GLUCOSE~\cite{mostafazadeh2020glucose}, ASER~\cite{zhang2020aser, zhang2022aser}, and CausalBank~\cite{ijcai2020guided}.
For the Last Letter Connection task in Symbolic Reasoning, we manually construct a dictionary KB for each word in Wiktionary. For example, the triple of the word ``system'' is ``(system, last letter, m)''.
For the remaining datasets (e.g., arithmetic reasoning and coin dataset), we do not perform F\textsuperscript{2}-Verification because we can not find any KBs for these tasks.
More details is shown in Appendix~\ref{app:implementation_details}.

% \begin{itemize}
%     \item \textbf{OPT-\sqs{Param}B}~\cite{zhang2022opt}, which is an open-sourced large causal language model with varying scales. \sqs{details TBD}
%     \item \textbf{GPT-3}~\cite{Brown2020Language}, specifically, the public engine \textit{text-davinci-003} with 175 billion parameters is employ.
%     \item \textbf{PaLM-540B}~\cite{Aakanksha2022PaLM}, which is pre-trained on a diverse range of curated sources ({e.g.,} books, Wikipedia, and codes) with decoder-only architecture.
% \end{itemize}

% sqs: OPT-?B, davinci-175B, and PaLM-540B
% sqs: OPT scale unkown, @wjn, use this doc: https://github.com/facebookresearch/metaseq/blob/main/projects/OPT/chronicles/README.md

% sqs: add CoT, ZS-CoT, and AutoCoT as baselines, subject to change
\paragraph{Baselines.}

In our experiments, 
we first consider few-shot/zero-shot standard prompting (SP) popularized by~\citet{Brown2020Language} as the naive baselines, and then some prevailing methods serve as strong baselines.
% 1) \textbf{Chain-of-thought (CoT)}~\cite{Wei2022Chain}: standard chain-of-thought prompting that provides four to eight hand-crafted examples, encompassing a sequence of intermediate reasoning steps.
% 2) \textbf{Zero-shot-CoT}~\cite{Kojima2022Large}: a zero-shot template-based approach for CoT, which diverges from the original method as it eliminates the need for step-by-step few-shot examples through adding task-agnostic prompts.
% 3) \textbf{Auto-CoT}~\cite{Zhang2022Automatic}: constructing exemplars for CoT through selecting diverse questions to build illustrative demonstrations.
1) Chain-of-thought prompting (Few-Shot CoT \& Manual CoT)~\cite{Wei2022Chain}, 2) Zero-Shot CoT~\cite{Kojima2022Large}, 3) Auto-CoT~\cite{Zhang2022Automatic}, 4) Complexity-based prompting (ComplexCoT)~\cite{fu2023complexcot}. We also integrate Self-Consistency (SC) into Manual CoT, ComplexCoT, and our CoK when validating the~\emph{gpt-3.5-turbo} model. The number of sampled reasoning paths is 10.

\begin{figure*}
%\vspace{-.25em}
\centering
\begin{tabular}{cc}
\begin{minipage}[t]{0.49\linewidth}
    \includegraphics[width = \linewidth]{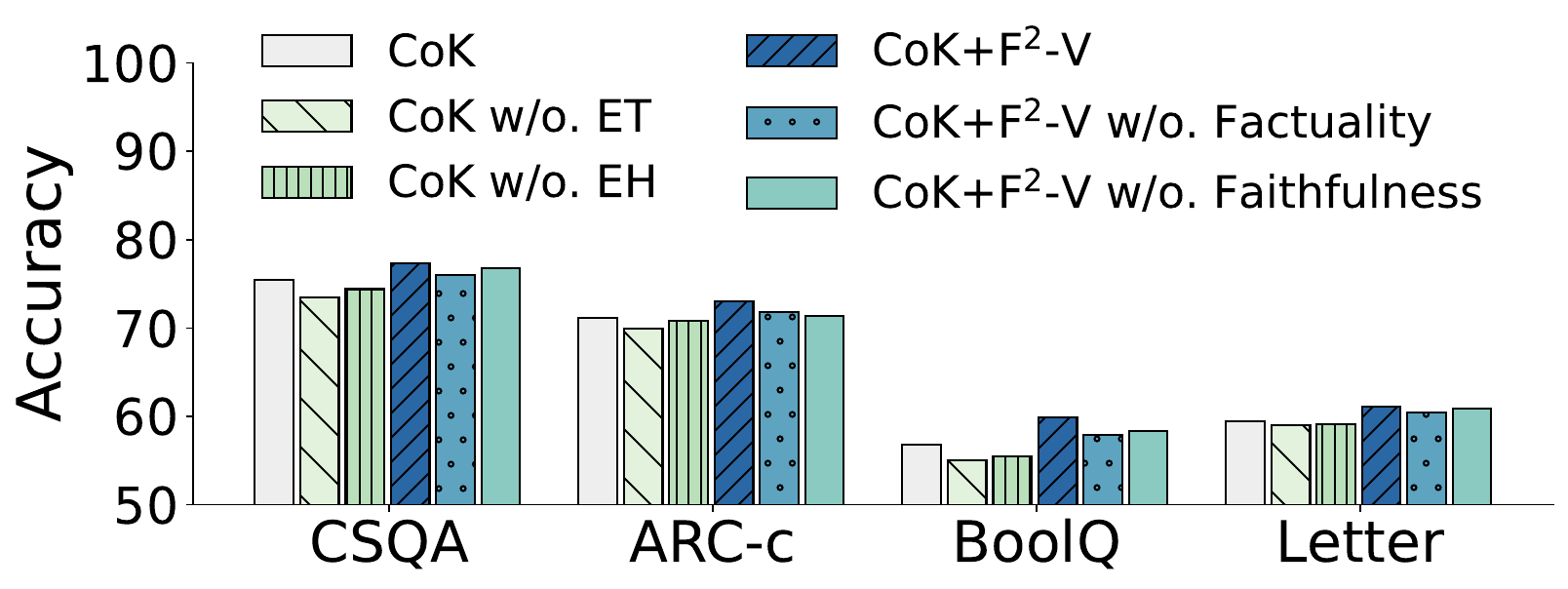}
\caption{
Ablation study results: accuracy when we remove different components.
}
\label{fig:ablation}
\end{minipage}
\begin{minipage}[t]{0.49\linewidth}
    \includegraphics[width = \linewidth]{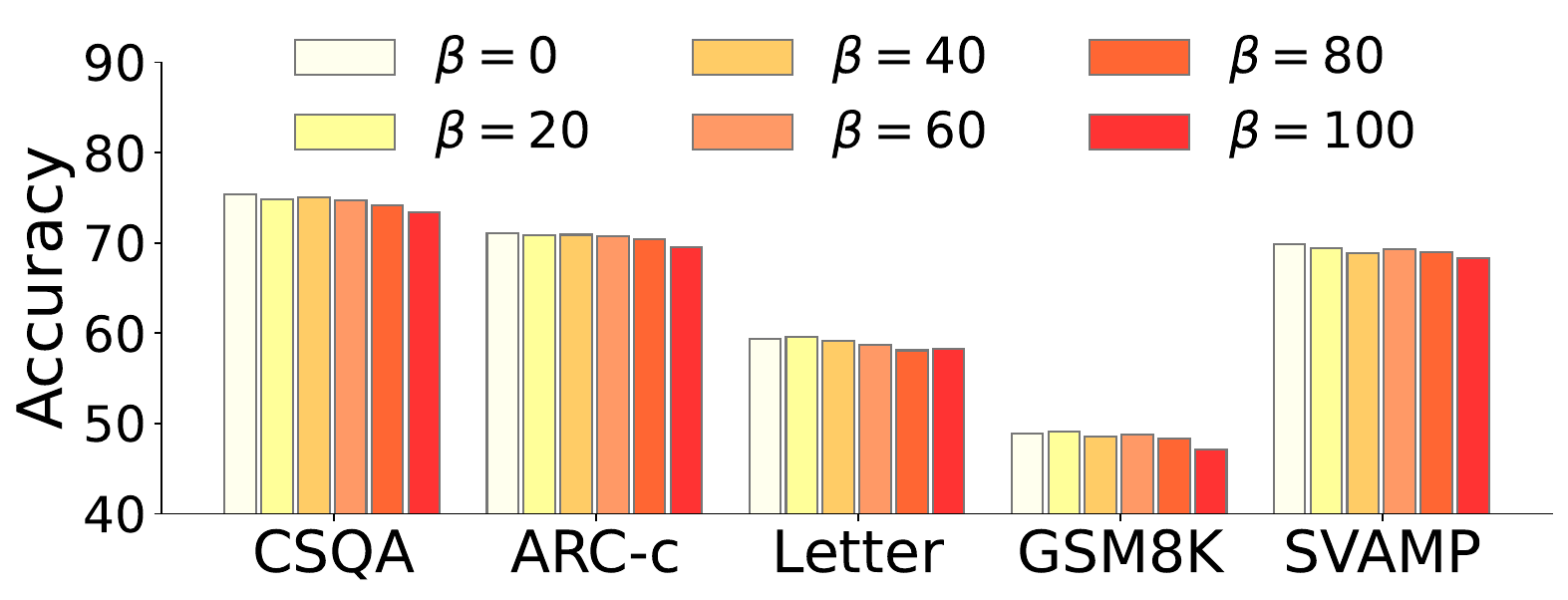}
\caption{
Effect of wrong demonstrations with $\beta$\% wrong evidence triples.
}
\label{fig:wrong_evidence_triples}
\end{minipage}
\end{tabular}
\vspace{-1.em}
\label{fig:preexp1}
%\vspace{-.5em}
\end{figure*}

\subsection{Competitive Performance of CoK}
\label{sec:main_results}

% \paragraph{Commonsense \& Factual Reasoning.}

We first evaluate on commonsense and factual reasoning. As shown in Table~\ref{tab:main_results}, we make the following observations:
1) CoK prompting steadily exceeds the performance of the previous CoT strategies.
% consistently matches or exceeds
Specifically, our method respectively achieves 1.9\%, 1.2\%, 0.9\%, 1.5\%, 0.8\%, and 1.8\% improvement with~\emph{text-davinvi-002}, and respectively achieves 0.6\%, 1.2\%, 0.9\%, 0.8\%, 1.9\%, and 2.8\% improvement with~\emph{gpt-3.5-turbo}. This demonstrates that the combination of explicit evidence triples and explanation can boost the LLMs' reasoning ability.
This also suggests that a better elicitation format is critical for prompt-based learning.
2) Based on F\textsuperscript{2}-Verification, the performances can be further enhanced across tasks.
In particular, the performance of CoK + F\textsuperscript{2}-Verification nearly approaches fine-tuning on the StrategyQA and ARC-c tasks.
This indicates that conducting post-verification and correcting false evidence triples by injecting ground-truth knowledge is crucial to the reasoning.
% 3) CoK steadily outperforms the ComplexCoT method by a significant margin, which requires a much higher computational cost than our approach.
3) CoK steadily outperforms the ComplexCoT, which requires a much higher computational cost than our, by a significant margin.
% 3) For the CSQA and OpenBookQA tasks, few-shot SP occasionally obtains the best results, we think that the model may be stuck in a thinking dilemma on some samples. Despite this,
% our method can still outperform other strong baselines.

% \paragraph{Symbolic \& Arithmetic Reasoning.}
We also explore how can CoK prompting adapt to non-knowledge-intensive tasks, such as symbolic and arithmetic reasoning.
% which are the most challenging tasks for LLMs.
% For symbolic reasoning, we choose the last letter connection and coin flip tasks.
Results in Table~\ref{tab:main_results} suggest that CoK prompting can also make high improvements on these tasks,
% For arithmetic reasoning tasks, it is hard to find an off-the-shelf KB to estimate the evidence triples.
% Nonetheless, we still achieve the best accuracies in Table~\ref{tab:main_results}, 
indicating that decomposing the reasoning chains into explicit triples is helpful for LLMs to understand complex tasks.

Finally,
we compare some ensemble baselines with~\emph{self-consistency} (SC), and we find 1) self-consistency can substantially improve the accuracy on Manual CoT, ComplexCoT and CoK, 
2) CoK + SC + F\textsuperscript{2}-V achieve the best performances on most tasks, where SC boosts reasoning by assembling all reasoning paths at each rethinking iteration and F\textsuperscript{2}-V boosts reasoning by assembling all iterations.
% \sqs{models}
% To evaluate the generalization ability of CoK using different LLMs.

\subsection{Discussions}
\paragraph{Ablation Study.}
\label{sec:discussions}

% \begin{wrapfigure}{r}{0.55\textwidth}
% % \vspace{-4mm}
% \includegraphics[width=\linewidth]{images/ablation}
% \caption{
% Ablation study results: accuracy when we remove different components.
% }
% \label{fig:ablation}
% % \vspace{-5pt}
% \end{wrapfigure}

In this section, we aim to explore how much each part of the component contributes to the performance.
We perform an ablation study to see how the performance changes.
% Thus, we design six variants: 1) \textbf{CoK} is the basic prompt without the rethinking process; 2) \textbf{CoK w/o. ET} consists of only explanation hints (is the same as CoT); 3) \textbf{CoK w/o. EH} only has evidence triples; 4) \textbf{CoK + F\textsuperscript{2}-V} denotes the whole framework; 5) \textbf{CoK + F\textsuperscript{2}-V w/o. Factuality} aims to only use faithfulness score in the rethinking process; and 6) \textbf{CoK + F\textsuperscript{2}-V w/o. Faithfulness} aims to only use a factuality score. 
We conduct the experiments on four tasks, including CSQA, ARC-c, BoolQ, and the Last Letter Connection.
The ablation settings are shown in Appendix~\ref{app:ablation_settings}.
% \begin{wrapfigure}{r}{0.55\textwidth}
% \vspace{-25pt}
% \includegraphics[width=\linewidth]{images/wrong_et}
% \caption{
% Effect of wrong demonstrations with $\beta$\% wrong evidence triples.
% }
% \label{fig:wrong_evidence_triples}
% \vspace{-10pt}
% \end{wrapfigure}
Results in Figure~\ref{fig:ablation} demonstrate that the performance drops when removing each component, which shows the significance of all components.
For CoK, we can see that the performance of the variant CoK w/o. ET is lower than CoK w/o. EH on all the tasks, which suggests that urging the LLM to generate explicit evidence triples is the most important contribution to the performance.
In addition, both the evidence triples and explanation hints can be fully utilized in the rethinking process because they can guide the LLM to verify the reasoning chains via either factuality or faithfulness.

% \begin{table}[t]
% \centering
% \small
% \label{tab:irrelevant_demonstration}
% \resizebox{\linewidth}{!}{
% % \begin{small}
% \begin{tabular}{lcccc}
% \toprule
% Model & CSQA   & StrategyQA & OpenBookQA & ARC-c  \\
% \midrule
% CoK &  75.4  &  66.6  &  59.4  & 48.9  \\
% CoK (GSM8K) &  73.6  & 61.4 &  55.0  & 47.3 \\
% \bottomrule
% \end{tabular}
% % \end{small}
% \caption{Effect of wrong demonstrations with irrelevant demonstrations.
% }
% }
% \end{table}

\begin{table}[t]
\centering
% \small
\resizebox{\linewidth}{!}{
\begin{tabular}{lcccc}
\toprule
Domain $d$ & CSQA   & StrategyQA & OpenBookQA & ARC-c \\
\midrule
$\#d\rightarrow \#d$ &  75.4  &  66.6  &  59.4  & 48.9  \\
GSM8K$\rightarrow \#d$ &  73.6  & 61.4 &  55.0  & 47.3 \\
\bottomrule
    \end{tabular}
    }
\caption{Domain adaptation results (\%). $\#d$ means the domain of exemplars, $\#d\rightarrow \#d$ means the examples sampled from the current domain, while GSM8K$\rightarrow \#d$ means the current task use GSM8K exemplars.}
\label{tab:irrelevant_demonstration}
\vspace{-1em}
\end{table}

\paragraph{Effect of Wrong Demonstrations.} 
Recall the discussion in Section~\ref{sec:exemplars_construction} about the demonstrations that may have some mistakes.
To see if chain-of-knowledge prompting has a similar phenomenon to previous works~\cite{Wang2022Towards} that there is no strong connection between the validity of reasoning chains and the performance of model prediction.
We perform negative random replacement when constructing exemplars. Specifically, we choose $\beta$\% evidence triples in each in-context example and replace them randomly from the KB to form a wrong reasoning path.
% We design two settings: 1) $\beta$\% error rate of evidence triples: for each exemplar, we choose $\beta$\% evidence triples and replace them randomly from the KB.
% 2) Irrelevant demonstrations: for each exemplar, we replace the prompt with an alternative exemplar.
% Specifically, we choose CSQA, StrategyQA, OpenBookQA, and ARC-c tasks from commonsense reasoning, and for each task, we select the demonstrations from the GSM8K task, which is very different from them.
We choose five tasks and draw some bar charts in Figure~\ref{fig:wrong_evidence_triples}.
Results illustrate that the accuracy rate drops slightly as the value of $\beta$ increases from $0$ to $100$.
However, even if the evidence triplets are all wrong, the performance will not decrease significantly. This phenomenon is counter-intuitive, yet, in line with the expected situation we considered before.

\paragraph{Domain Adaptation of Demonstrations.}

To investigate the adaptation of CoK prompting, for each exemplar, we replace the prompt with an alternative exemplar from other domains.
% We investigate the domain adaptation of CoK prompting in Table~\ref{tab:irrelevant_demonstration}. 
Specifically, we choose CSQA, StrategyQA, OpenBookQA, and ARC-c tasks from commonsense reasoning, and for each task, we select the demonstrations from the GSM8K task, which is very different from them.
The results for the domain adaptation settings are shown in Table~\ref{tab:irrelevant_demonstration}.
% The performance is exciting because we find 
We find it exciting because the LLM can easily know how to follow the chain-of-knowledge paradigm to solve a new problem, although the given prompt is completely irrelevant.

\begin{figure}[t]
\includegraphics[width=\linewidth]{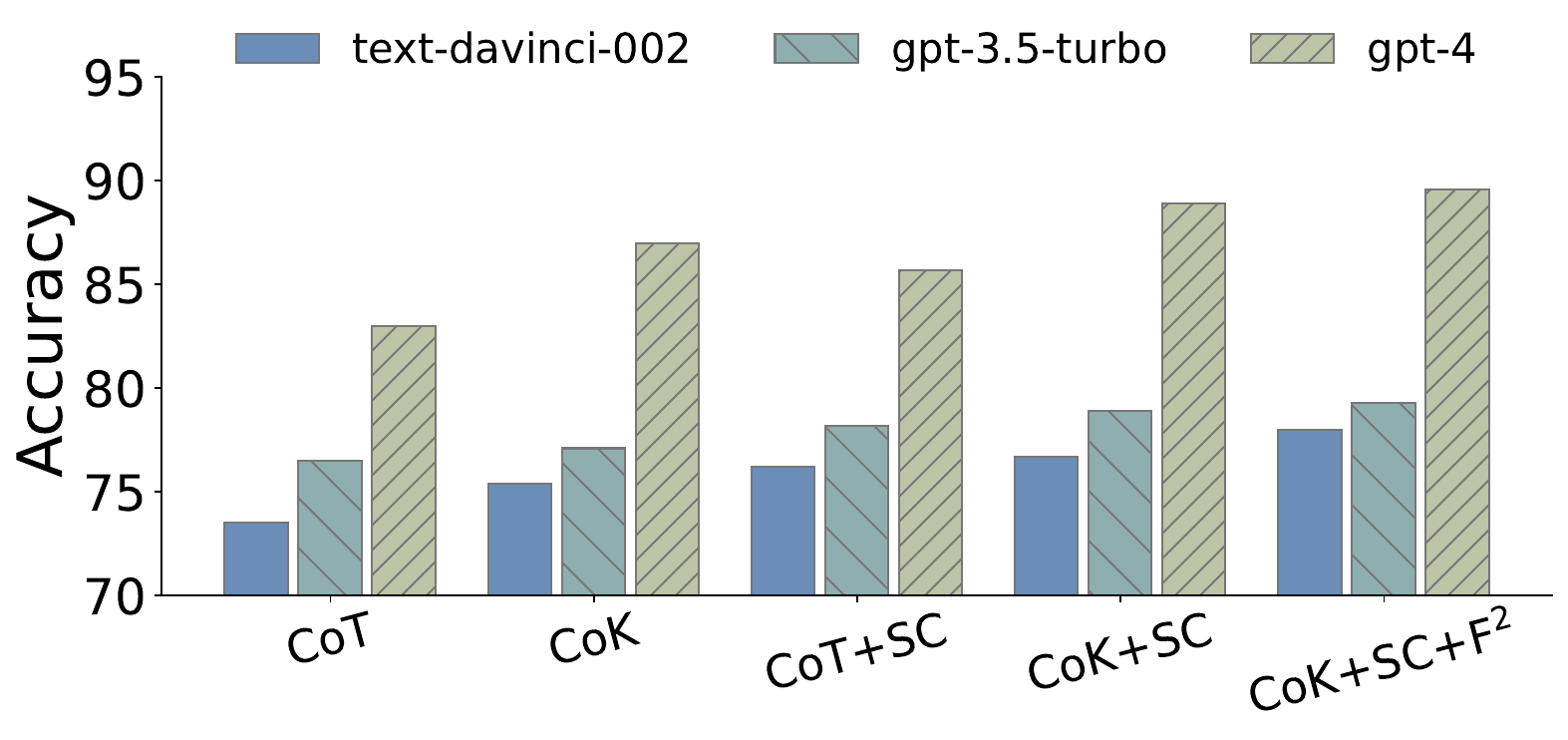}
	\caption{Comparison of CoT, CoT+SC, CoK, CoK+SC and CoK+SC+F\textsuperscript{2}-V over CSQA when using different backbone.}
 \label{fig:different_backbone_csqa}
\end{figure}

\paragraph{Model Effectiveness}
To investigate the effectiveness of CoK when applying different backbones, we extra choose~\emph{gpt-4} to evaluate CSQA and GSM8K. As shown in Figure~\ref{fig:different_backbone_csqa}, by comparing with CoT and CoT+SC, CoK and CoK+SC achieve average performance improvements of 1.6\% on~\emph{text-davinci-002}, 1.4\% on~\emph{gpt-3.5-turbo} and 1.0\%
on~\emph{gpt-4}, indicating that CoK is adaptable to various LLMs and effectively
boosts performance across different backbones.

\begin{figure}[t]
\centering
\includegraphics[width=0.9\linewidth]{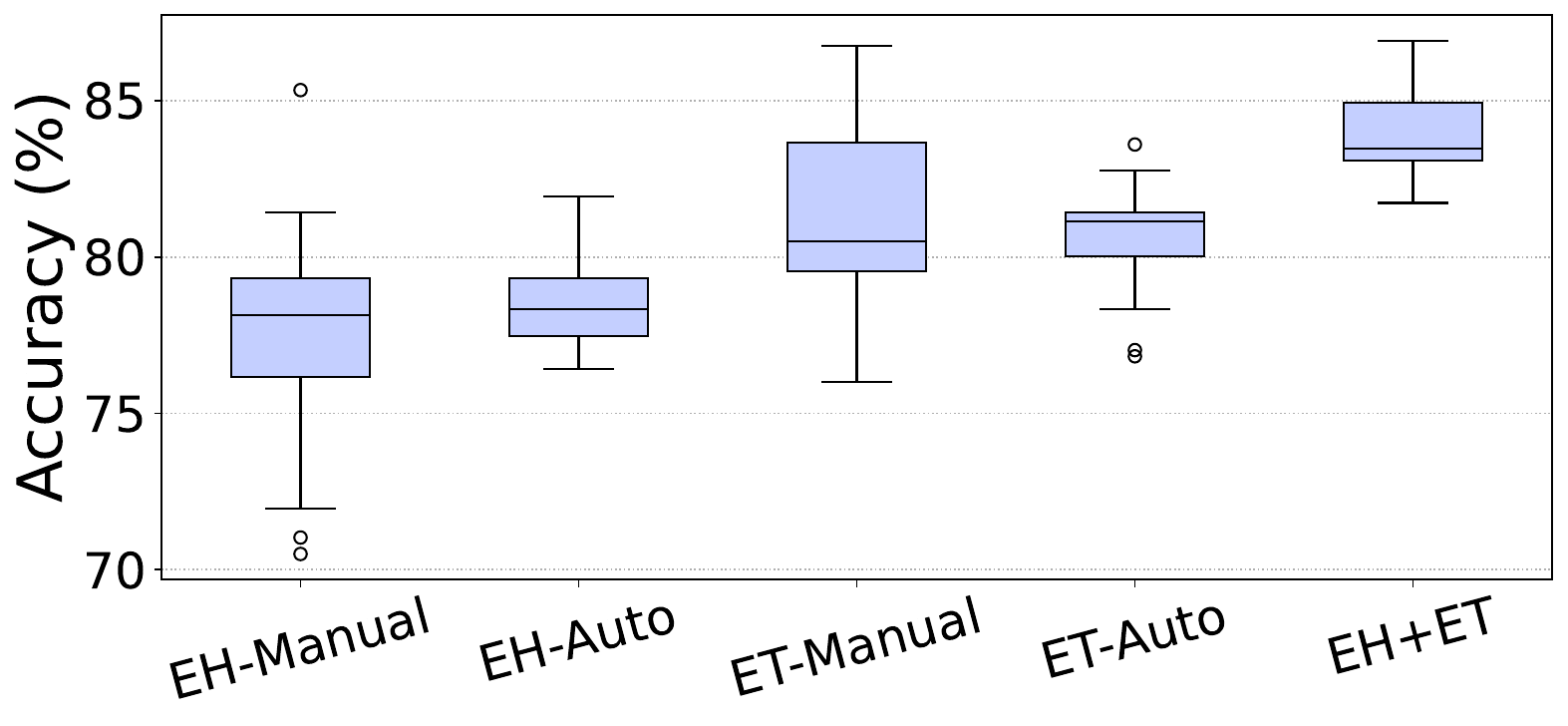}
	\caption{Performance of~\emph{gpt-3.5-turbo} over GSM8K with different prompt.}
 \label{fig:prompt_effectiveness}
\end{figure}

\begin{figure}[t]
\centering
\includegraphics[width=\linewidth]{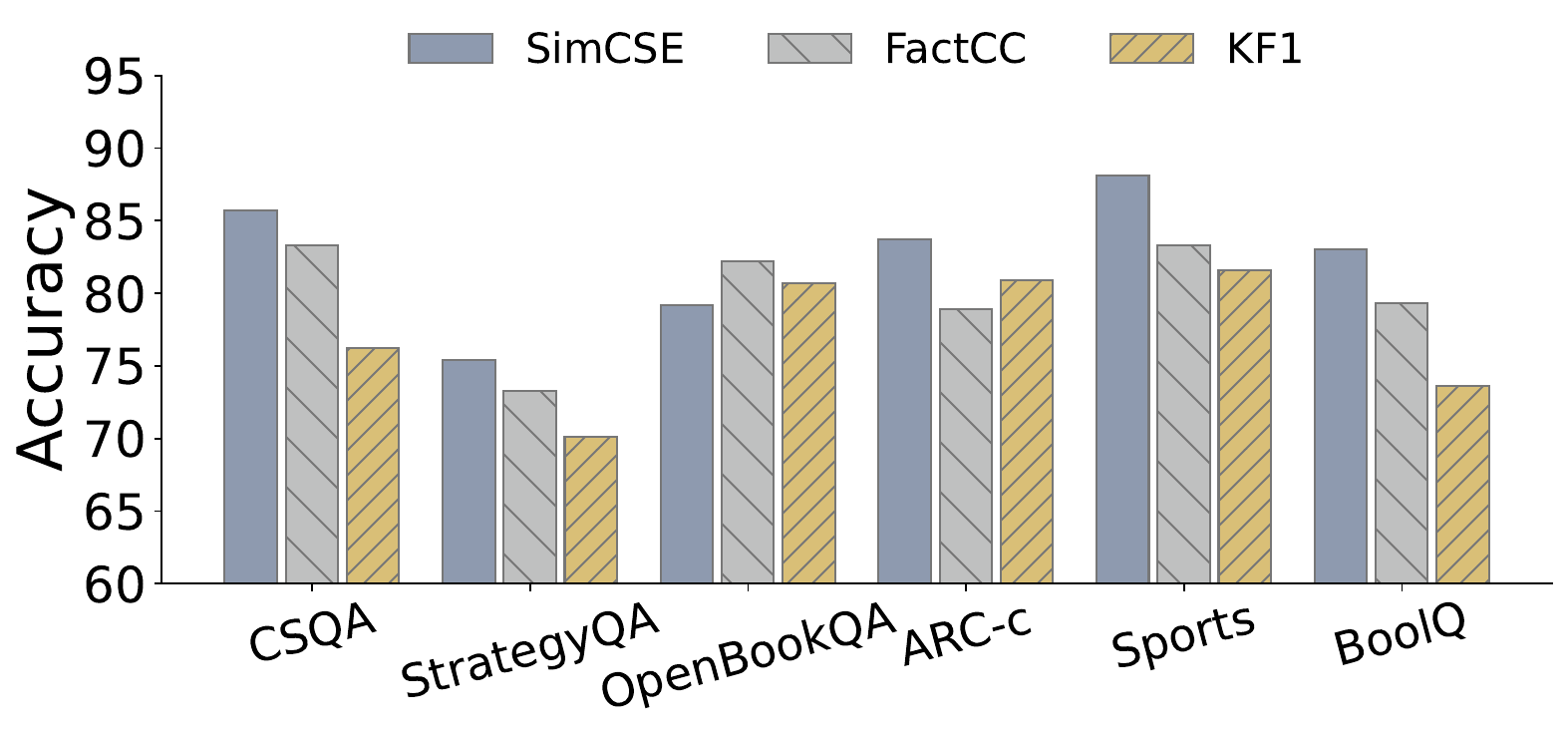}
	\caption{Hallucination Evaluation of different faithfulness score $f_u(\cdot)$.}
 \label{fig:different_score}
\end{figure}

\begin{figure}[t]
\centering
\begin{tabular}{cc}
\begin{minipage}[t]{0.5\linewidth}
    \includegraphics[width = 0.9\linewidth]{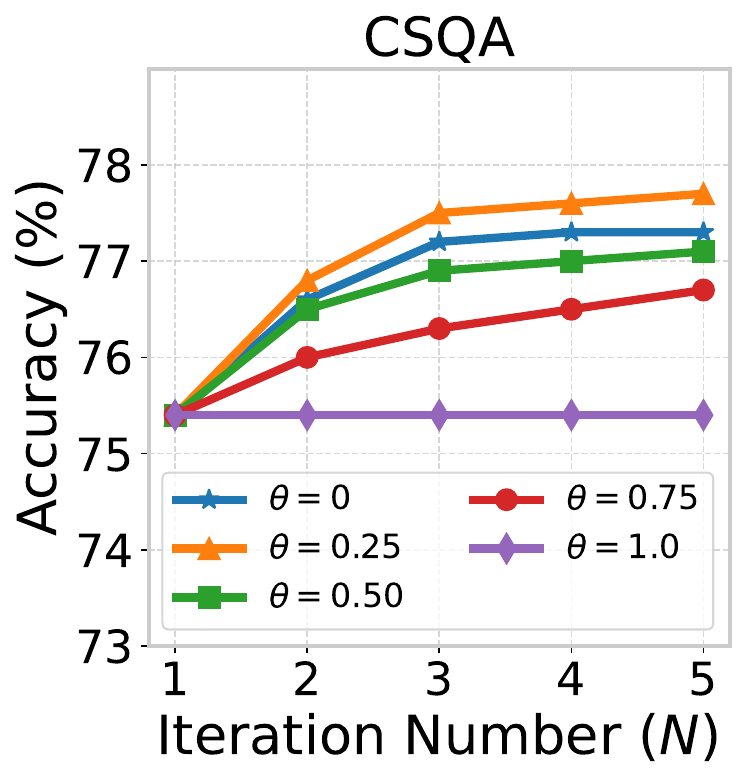}
\end{minipage}
\begin{minipage}[t]{0.5\linewidth}
    \includegraphics[width = 0.9\linewidth]{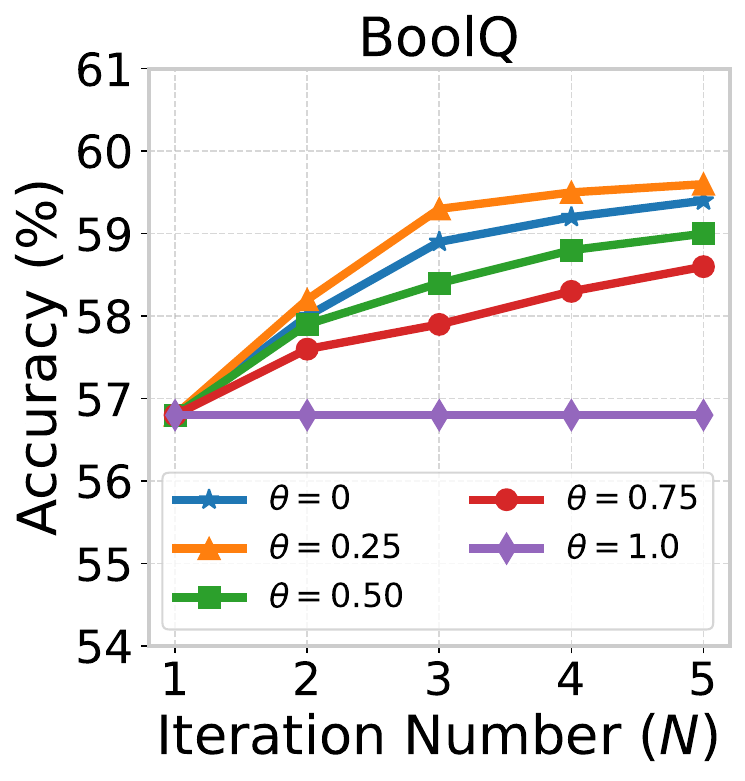}
\end{minipage}
% \begin{minipage}[t]{0.33\linewidth}
%     \includegraphics[width = 0.85\linewidth]{images/rethinking_3}
% \end{minipage}
\end{tabular}
\vspace{-.75em}
\caption{Effectiveness of different rethinking iteration $N$ and reliability threshold $\theta$ on \emph{text-davinci-002}.}
\label{fig:rethinking_effect}
\end{figure}

\paragraph{Prompt Engineering.}

In Figure~\ref{fig:prompt_effectiveness}, we analyze the effectiveness of different prompt engineering strategies. ``Manual'' denotes constructing prompt via human annotation, while ``Auto'' means to use zero-shot CoT and KB to build prompt.
Results demonstrate that leveraging zero-shot CoT with KB can reduce the variance and improve the accuracy.

\paragraph{Hallucination Evaluation.}

To investigate the hallucination, we choose different faithfulness scores $f_u(\cdot)$ to make the comparison. Apart from SimCSE, we also choose FactCC~\cite{Kryscinski2020Evaluating} and Knowledge F1 (KF1)~\cite{Shuster2021Retrieval}.
As shown in Figure~\ref{fig:different_score}, we choose all tasks from commonsense reasoning to make an evaluation. We find that our framework can achieve the highest accuracy on most of the tasks when using SimCSE as a faithfulness score, which indicates the effectiveness of hallucination reduction.

\paragraph{Effectiveness of Rethinking Process.} 
% 为了考虑预算问题，我们只选择3个数据集，且每个数据集只挑选100个样本进行实验，N的值设置为10
% 一共展示3个图，每个图代表一个数据集，每个图中有若干曲线，每个曲线表示选择某个reliability阈值条件下，rethinking过程中的准确率情况
% 每个坐标点，表示当迭代最大次数N为x，阈值为y时的准确率。
Recall the \emph{rethinking process}, when the reasoning chains generated by the LLMs fail to pass verifications and the reliability score is below the threshold $\theta$, 
we provide them with additional opportunities to regenerate in the rethinking stage. 
% Here, we set different $\theta$ and $N$ to examine their influence on task performance.
Figure~\ref{fig:rethinking_effect} demonstrates the effectiveness over three tasks with different combinations of rethinking iteration number $N\in\{1, 2, 3, 4, 5\}$ and threshold $\theta\in\{0, 0.25, 0.5, 0.75, 1.0\}$.
From the analysis,
we can draw some following suggestions.
1) In most cases, the accuracy increases a lot when the LLM rethinks step-by-step in the first 3 iterations.
2) The rethinking process converges faster when using a smaller threshold.
It is not difficult to understand that when the threshold is small, almost all testing queries will be injected with knowledge and regenerated, which is similar to the method of combining \emph{self-consistency} and F\textsuperscript{2}-Verification.
Interestingly, we observe that the performance may decrease by about 2\% when $\theta<0.25$, we blame it on an over-injection problem because it may inject some irrelevant or inconsistent information.

\section{Conclusion}

% simple yet effective
% \sqs{conclusion: lxnb} 
We propose chain-of-knowledge prompting, which aims to decompose the reasoning chains derived from the LLMs into multiple evidence triples and explanation hints, to further improve the reasoning capabilities.
Based on the chain-of-knowledge prompt, we introduce F\textsuperscript{2}-Verification and fully exploit external knowledge bases to perform post-verification for the generated reasoning chains in terms of factuality and faithfulness.
A rethinking process then be used to inject knowledge to correct the false evidence triples and elicit the LLM to regenerate the answer.
Our extensive results show that it outperforms other prompt methods over multiple reasoning tasks.
In the future, we will 1) further improve the performance of other scale LLMs, 2) extend the KB to search engines to realize real-time verification, and 3) perform interpretability analysis on LLMs' reasoning.

% Through a wide range of experiments on \sqs{tasks}, 
% we demonstrate that CoK leads to superior performance while ensuring factuality and faithfulness.
% In addition to enhancing task performance, 
% CoK can also help mitigate issues such as hallucination within the model,
% preventing the model from generating unfactual and unfaithful information. 
% We believe that it will have a positive impact on the NLP community.

% sqs: add limitations and broader impact
% 1. kb might not cover all questions
% 2. rethink -> more API calls
% ethics: external KBs will not cause biases

% \begin{ack}
% Use unnumbered first level headings for the acknowledgments. All acknowledgments
% More information about this disclosure can be found at: \url{https://neurips.cc/Conferences/2023/PaperInformation/FundingDisclosure}.
% Do {\bf not} include this section in the anonymized submission, only in the final paper. You can use the \texttt{ack} environment provided in the style file to autmoatically hide this section in the anonymized submission.
% \end{ack}

\section*{Limitations}

Our work is based on prompting methods for large language models and achieves outstanding performance across several benchmarks. However, it still carries the following limitations:
(1) The evidence triples in knowledge bases are finite, 
which might not ensure comprehensive coverage of the model's requirements for all questions.
% Additionally, 
(2) In light of the integration of the re-thinking algorithm, 
CoK might require more API calls compared to vanilla CoT methods.

\section*{Social Impact and Ethics}

In terms of social impact,
the knowledge bases we utilize are all from publicly available data sources. 
Infusing factual knowledge into the model's reasoning process will not introduce additional bias. 
Moreover, it can to some extent prevent the model from providing irresponsible and harmful answers.

\section*{Acknowledgement}
This work is supported by National Natural Science Foundation of China under Grant No. 62202172, No. 62377012, and No. U1911203, and Shanghai Science and Technology Committee General Program No. 22ZR1419900.
We thank Nuo Chen for the discussion and valuable feedback.

\bibstyle{acl}
\bibliography{custom}

% \newpage

\appendix

\begin{table*}
    \centering
    \scriptsize
    \vspace{5pt}
    \resizebox{\linewidth}{!}{
    \begin{tabular}{p{2.5cm} | p{4.8cm} | p{3cm} | p{4.8cm} }
    \toprule
        \textbf{Query} & \textbf{Evidences Triples \& Explanations} & \textbf{Injected Knowledge Prompt} & \textbf{Rethinking}  \\
        \midrule
        \textbf{[$K$ exemplars]}$\backslash$n$\backslash$n What island country is ferret popular? Answer Choices: (A) own home (B) north carolina (C) great britain (D) hutch (E) outdoors
        & A: Evidence triples:
        1. (ferret, isA, animal)$\backslash$n
        2. (country, isA, place)$\backslash$n
        3. (ferret, popular, Britain)$\backslash$n
        \textcolor{red}{4. (north carolina, located in, Britain)}$\backslash$n
        Explanation hints: The answer must be a place where ferrets are popular. Of the above choices, \textcolor{red}{only North Carolina is a place where ferrets are popular}.$\backslash$n
        \textcolor{red}{So the answer is (D)}.
        & \textbf{[$K$ exemplars]}$\backslash$n$\backslash$n What island country is ferret popular? Answer Choices: (A) own home (B) north carolina (C) great britain (D) hutch (E) outdoors$\backslash$n
        \textcolor{blue}{(North Carolina, located in, US.)$\backslash$n(Great Britain, located in, Britain)}
        & A: Evidence triples:
        1. (ferret, isA, animal)$\backslash$n
        2. (country, isA, place)$\backslash$n
        3. (ferret, popular, Great Britain)$\backslash$n
        \textcolor{green}{4. (Great Britain, located in, Britain)}$\backslash$n
        Explanation hints: The answer must be a place where ferrets are popular. Of the above choices, \textcolor{green}{only North Carolina is a place where ferrets are popular}.$\backslash$n
        \textcolor{green}{So the answer is (C)}.  \\
        \midrule
         \textbf{[$K$ exemplars]}$\backslash$n$\backslash$n
         Take the last letters of each words in ``Prince Rene Vishal Patrick'' and concatenate them.
         & A: Evidence triples:$\backslash$n1. (Prince, last latter, e) $\backslash$n2. (Rene, last latter, e) $\backslash$n\textcolor{red}{3. (Vishal, last latter, i)} $\backslash$n4. (Patrick, last latter, k) $\backslash$n5. (final answer, is, eelk) $\backslash$n Explanation hints: The last letter of ``Prince'' is ``e''. The last letter of ``Rene'' is ``e''. \textcolor{red}{The last letter of ``Vishal'' is ``i''.} The last letter of ``Patrick'' is ``k''. \textcolor{red}{Concatenating them is ``eeik''. $\backslash$n So the answer is eeik}.  
         & \textbf{[$K$ exemplars]}$\backslash$n$\backslash$n
         Take the last letters of each words in ``Prince Rene Vishal Patrick'' and concatenate them.$\backslash$n
         \textcolor{blue}{(Vishal, last latter, l)}
         & A: Evidence triples:$\backslash$n1. (Prince, last latter, e) $\backslash$n2. (Rene, last latter, e) $\backslash$n\textcolor{green}{3. (Vishal, last latter, l)} $\backslash$n4. (Patrick, last latter, k) $\backslash$n5. (final answer, is, eelk) $\backslash$n Explanation hints: The last letter of ``Prince'' is ``e''. The last letter of ``Rene'' is ``e''. \textcolor{green}{The last letter of ``Vishal'' is ``l''.} The last letter of ``Patrick'' is ``k''. \textcolor{green}{Concatenating them is ``eelk''. $\backslash$n So the answer is eelk}.    \\
         \bottomrule
    \end{tabular}
    }
    \caption{Case study on CSQA and Last Letter Connection: the chain-of-knowledge prompting with the rethinking process. The tokens in red, blue, and green respectively denote the wrong rationales, the injected knowledge, and the corrected rationales.}
    \label{tab:case}
\end{table*}

\section{Case Study}

We end this section with a case study to show the effectiveness of our proposed chain-of-knowledge prompting and the rethinking process with F\textsuperscript{2}-Verification.
We randomly choose two examples from CSQA and Last Letter Connection tasks, and the results are listed in Table~\ref{tab:case}.
We can see that our proposed method can effectively generate explicit evidence triples with corresponding explanation hints, and the wrong triples can be detected through the proposed F\textsuperscript{2}-Verification.
During the rethinking process, the LLM can be guided with a new prompt with injected knowledge and then reach the correct answer.
We also find the LLM can spontaneously check whether the injected knowledge is useful for the question solution (as shown in Appendix~\ref{app:case_study}), which indicates that our method can achieve meticulous injection and avoid injecting useless knowledge.

\section{Statistics of Datasets}

We evaluate CoK on 12 publicly available benchmark datasets that cover arithmetic reasoning, commonsense reasoning, symbolic reasoning and natural language understanding tasks. The statistics of the datasets are shown in Table~\ref{tab:dataset_stats}.

% sqs: 因为有NLU任务，我去掉了前面文章里“Average words”这一个属性的描述
\begin{table*}[htb]
\centering
% \small

\setlength{\tabcolsep}{13.6pt}

\resizebox{\linewidth}{!}{
\begin{tabular}
{lcccccc}
\toprule
Dataset & Number of samples &Average words &Answer Format & Licence\\
\midrule
{CSQA} & 1,221 & 27.8 & Multi-choice & Unspecified \\
{StrategyQA} & 2,290 & 9.6  & Yes or No  &Apache-2.0 \\
{OpenBookQA} & 500  & 27.6  & Multi-choice & Unspecified  \\
{ARC-c} & 1,172  & 47.5  & Multi-choice  & CC BY SA-4.0  \\
{Sports} & 1,000 & 7.0  & Yes or No  & Apache-2.0 \\
{BoolQ} &  3,270 & 8.7 & Yes or No  &  CC BY SA-3.0 \\
{Last Letters} & 500 & 15.0 & String & Unspecified \\
{Coin Flip}  & 500 & 37.0 & Yes or No & Unspecified \\
{GSM8K}  & 1,319 & 46.9 & Number & MIT License \\
{SVAMP}  & 1,000 & 31.8 & Number & MIT License \\
{AQuA} & 254  & 51.9 & Multi-choice  &  Apache-2.0 \\
{MultiArith} & 600  & 31.8 & Number & CC BY SA-4.0  \\
\bottomrule
\end{tabular}
}
\caption{Dataset Descriptions.}
\label{tab:dataset_stats}
\end{table*}

The detailed information of each dataset is shown in the follow:
\paragraph{Commonsense \& Factual reasoning}
\begin{itemize}
    \item CSQA \cite{talmor2019commonsenseqa}:
    it is a commonsense QA task and the answer format is multi-choice question answering. The homepage is \url{https://www.tau-nlp.org/commonsenseqa}, and \url{https://github.com/jonathanherzig/commonsenseqa}.
    
    \item StrategyQA \cite{geva2021strategyqa}: it is a commonsense QA task with Yes or No answer format. We use the open-domain setting (question-only set) from  \cite{srivastava2022imitation}: \url{https://github.com/google/BIG-bench/tree/main/bigbench/benchmark_tasks/strategyqa}.
    The original dataset is from \url{https://github.com/eladsegal/strategyqa}, MIT license: \url{https://github.com/eladsegal/strategyqa/blob/main/LICENSE}.

    \item OpenBookQA
    \cite{mihaylov2018openbookqa}: it is a multi-choice QA task to evaluate commonsense knowledge. The original dataset is from \url{https://allenai.org/data/open-book-qa}.

    \item ARC-c
    \cite{clark2018think}: it is a multip-choice commonsense QA task.
    The original dataset is from \url{https://allenai.org/data/arc}. CC BY SA-4.0 license: \url{https://creativecommons.org/licenses/by-sa/4.0/}.
    
    \item Sports understanding from BIG-Bench \cite{srivastava2022imitation}: the answer format is Yes or No. Apache License v.2: \url{https://github.com/google/BIG-bench/blob/main/LICENSE}.

    \item BoolQ
    \cite{clark2019boolq}: it is a knowledge-intensive task and the format is Yes or No.
    The original dataset is from \url{https://github.com/google-research-datasets/boolean-questions}. CC BY SA-3.0 license: \url{https://creativecommons.org/licenses/by-sa/3.0/}.

\end{itemize}

\paragraph{Symbolic \& Arithmetic reasoning}
\begin{itemize}
    \item Last Letters \& Coin Flip
    \cite{Wei2022Chain} are novel benchmarks to evaluate whether the LLM can solve a simple symbolic reasoning problem.
    The last letters dataset is from \url{https://huggingface.co/datasets/ChilleD/LastLetterConcat}. The coin flip dataset is from \url{https://huggingface.co/datasets/skrishna/coin_flip}.

    \item GSM8K \cite{cobbe2021training}: \url{https://github.com/openai/grade-school-math}, MIT license: \url{https://github.com/openai/grade-school-math/blob/master/LICENSE}.

    \item SVAMP \cite{patel2021nlp}: \url{https://github.com/arkilpatel/SVAMP}, MIT license: \url{https://github.com/arkilpatel/SVAMP/blob/main/LICENSE}.
    
    \item AQuA \cite{Ling2017Program}: \url{https://github.com/deepmind/AQuA}, license: \url{https://github.com/deepmind/AQuA/blob/master/LICENSE}.

    \item Math Word Problem Repository 
    MultiArith \cite{Roy2015Solving}, license: CC BY 4.0, dataset: \url{https://huggingface.co/datasets/ChilleD/MultiArith}.
    
\end{itemize}

\begin{table*}[t]
\centering
% \vspace{-3mm}
\setlength{\tabcolsep}{1pt}
\vspace{2.8mm}
\resizebox{0.85\linewidth}{!}{
\begin{tabular}{p{0.28\textwidth}  p{0.7\textwidth}}
\toprule
% \cmidrule(r){1-2}
% \cmidrule(r){8-9}
Input   & Response \\
\midrule
\multicolumn{2}{l}{Vanilla: \emph{Knowledge Injection before Reasoning}} \\
\midrule
Q: Who plays mary jane in spiderman 3? $\backslash$n \textcolor{blue}{Knowledge: (Kirsten Dunst, (actors, jk simmons), (award received, Milos Foreman)), (Kirsten Dunst, actors, Kirsten Dunst, occupation, female model)} & 
The answer to the question is Kirsten Dunst. 
To explain our reasoning with the knowledge you provided:$\backslash$n - Kirsten Dunst played Mary Jane in Spiderman 3. To confirm this information, we can look at actors who were in Spiderman 3, and we know that J.K. Simmons was also in the film. \textcolor{red}{However, this information does not answer our question.}$\backslash$n - We can also look at awards received, and we know that Milos Forman received awards for a film he directed called "One Flew Over the Cuckoo's Nest" which Kirsten Dunst was not in.$\backslash$n - \textcolor{red}{We can look at Kirsten Dunst's occupation and know she is a female model, which does not help answer our question.}$\backslash$n - Therefore, we can conclude that Kirsten Dunst played Mary Jane in Spiderman 3. \\
\midrule
\multicolumn{2}{l}{Ours: \emph{Knowledge Injection after F\textsuperscript{2}-Verification}} \\
\midrule
Q: Who plays mary jane in spiderman 3? $\backslash$n \textcolor{blue}{Knowledge: (spiderman 3, stars, kirsten dunst)} & First, we need to recall which film we're talking about, which is Spiderman 3. $\backslash$n
Next, we need to think about the key players in the film, and who played Mary Jane in that particular movie. The actress who portrayed Mary Jane in Spiderman 3 is Kirsten Dunst. \\

\bottomrule
\end{tabular}
% \vspace{-3.6mm}
}
\caption{Performance of different knowledge injection. 
}
\label{tab:app_case}
\end{table*}

\section{Implementation Details}
\label{app:implementation_details}

\subsection{CoK Construction}
\label{app:cok_construction}

For each dataset, we aim to construct demonstrations with multiple well-designed exemplars. The prompt example of each dataset is shown in Appendix~\ref{app:cok_prompt_example}.

During the prompt construction, we first randomly select multiple labeled examples from the training set.
For a fair comparison, we directly choose the selected labeled data from~\cite{Wei2022Chain, Wang2022Self, Kojima2022Large}.
Specifically, we choose 8 labeled data for Coin Flip, ARC-c, AQuA, GSM8K, MUltiArith, CSQA, SVAMP, OpenBookQA; 4 labeled data for Last Letter Connection; 6 labeled data for Sports, BoolQ, StrategyQA.

For each label data, we first use zero-shot CoT~\cite{Kojima2022Large} to perform textual reasoning chain generation. We directly connect a simple prompt ``Let's think step by step.'' after the input query to elicit the LLM to generate rationale and the final answer~\footnote{To alleviate the generation bias problem, we also use the self-consistency decoding~\cite{Wang2022Self} to sample one rationale.}.
We then remove this prompt and rebuild the input query by concatenating the input query and the generated textual reasoning chain.

To construct evidence triples, we aim to retrieve some relevant knowledge triples from the pre-built knowledge base (as shown in Appendix~\ref{app:kb}).
During retrieval, given a textual reasoning chain (e.g., CoK-EH), we encode it with the basic sentence encoder model (e.g., BERT) and then retrieve the most relevant knowledge triple using the maximum inner product search tool SCaNN~\cite{Guo2020Accelerating}.
Due to the retrieved knowledge triples may consist of noises and redundant information. To improve the reliability of the evidence triples, we have invited five domain experts (including volunteer professors and Ph.D. students from diverse research areas) to manually annotate the evidence triples based on the retrieved knowledge triples.

To improve the annotation efficiency, we also employ the idea of self-training, which aims to generate annotated data based on very few data. 
Specifically, we can first manually annotate two labeled data with evidence triples and explanation hints to form a 2-shot CoK prompt. Then, each rest labeled data is concatenated with this 2-shot CoK prompt and the LLM can generate the corresponding rationale and answer.
Thus, we can invite these experts to verify and correct them.

Finally, we obtain five different annotated demonstrations. To select the best one for each dataset, before the self-training process, we randomly choose some examples from the training set to form a validation set and use it to perform an evaluation. The prompt which has the best accuracy value is chosen.

\subsection{Knowledge Bases}
\label{app:kb}

We follow~\citet{Pan2022Knowledge} to select six knowledge bases from different domains.
Our implementation closely aligns with the approach described by~\citet{wang2023hugnlp}.
\paragraph{Dictionary} We utilize lexical knowledge, which comprises definitions and example sentences of English words (e.g., \textit{< ``apple'', definition, ``A common, round fruit ...'' >}) from Wiktionary~\footnote{\url{https://en.wiktionary.org/wiki/Wiktionary:Main_Page}.}.
Specifically, we use the Wiktionary dump dated April 30, 2022 that contains 1.3M word definitions and 470K example sentences for 1M words/phrases.

\paragraph{Commonsense} We incorporate commonsense knowledge from ConceptNet~\cite{robyn2017conceptnet}, which encompasses a wide range of general daily life knowledge, for instance, \textit{< ``rabbit'', CapableOf, ``move fast'' >}.
We follow previous works~\cite{zhang2020aser} to include the core 600K high-quality triplets.

\paragraph{Causality} The external knowledge we utilize for causality is derived from the auto-extracted causal knowledge source, CausalBank~\cite{ijcai2020guided}. This database consists of large-scale English sentences that express cause-effect relationships, such as \textit{< ``babies cry'', because, ``they are lonely'' >}).

\paragraph{Entity} We integrate named entity knowledge in Wikidata5M~\cite{wang2021kepler}. For each entity, the properties are collected in the format of \textit{< ``Kepler space telescope'', Named after, ``Johannes Kepler'' >}).
Here, related sentences refer to the sentences from an entity’s own article, or the sentences of other articles that link to this entity.

\paragraph{Event} We take into account knowledge of everyday events, 
utilizing both human-constructed event knowledge graphs such as ATOMIC~\cite{sap2019atomic} and GLUCOSE~\cite{mostafazadeh2020glucose},
as well as automatically extracted event knowledge graphs like ASER~\cite{zhang2020aser, zhang2022aser}.

\paragraph{Scripts} Specifically, we use 325K triples that are in the form of < verbal information, context, nonverbal information >, where verbal information is an utterance, nonverbal information can be body movements, vocal tones, or facial expressions, etc., and context is the entire text of the scene from which the verbal-nonverbal pair is extracted.

\subsection{Ablation Settings}
\label{app:ablation_settings}

For the ablation study, we design six variants: 1) \textbf{CoK} is the basic prompt without the rethinking process; 2) \textbf{CoK w/o. ET} consists of only explanation hints (is the same as CoT); 3) \textbf{CoK w/o. EH} only has evidence triples; 4) \textbf{CoK + F\textsuperscript{2}-V} denotes the whole framework; 5) \textbf{CoK + F\textsuperscript{2}-V w/o. Factuality} aims to only use faithfulness score in the rethinking process; and 6) \textbf{CoK + F\textsuperscript{2}-V w/o. Faithfulness} aims to only use a factuality score. 

\section{Analysis}
% sqs: grammar correction. Why do verification is useful? -> Why is verification useful?
\subsection{Why is verification useful?}
\label{app:case_study}
We demonstrate some cases to show how the LLM verifies the usefulness of the injected knowledge.
We use \emph{gpt-3.5-turbo} model because it can solve a problem through conversation.
We define two following settings. 
1) Vanilla: knowledge injection before reasoning. We directly concatenate the related knowledge triples with the input query, and prompt the LLM to think step by step.
2) Ours: knowledge injection after F\textsuperscript{2}-Verification. We first use a CoK prompt to elicit the LLM to generate evidence triples and find error triples. Then, we correct them into corresponding ground truth triples and concatenate them with the input query.

As shown in Table~\ref{tab:app_case}, we can see that the LLM can spontaneously detect and analyze each injected triple.
For example, if the knowledge triple is useless or has no contribution to the reasoning, the LLM can talk to me about which and why are useless (the red text in Table~\ref{tab:app_case}).
This indicates that the knowledge provided by the traditional knowledge injection method is not completely useful.
In contrast, our approach can accurately locate the false reasoning evidence derived from the LLM after the verification stage, so as to perform targeted knowledge injection.

% \subsection{Effectiveness of KB}
% In this paper, we aim to leverage the knowledge from KB to enhance the reasoning ability. Alternatively, we can also use search engine

% \paragraph{Compared with Self-consistency.} 
% \paragraph{Compared with other repairs strategy.} 

% \section{Input/Output Examples}

% \input{tables/appendix-io-examples-letters}

% \input{tables/appendix-io-examples-coin}

\section{Exemplars with Chain-of-Knowledge Prompts}
\label{app:cok_prompt_example}

The details of our prompts are shown below.

\newpage

\begingroup
\begin{table}[H]
    \centering
    \small
    \begin{tabular}{p{\linewidth}}
        \toprule
        \textbf{Q:} Take the last letters of the words in "Elon Musk" and concatenate them. \\
        \textbf{A:} Evidence triples: \\
        1. (Elon, last latter, n) \\
        2. (Musk, last latter, k) \\
        3. (final answer, is, nk) \\
        Explanation hints: The last letter of "Elon" is "n". The last letter of "Musk" is "k". Concatenating them is "nk". \\
        So the answer is nk. \\
        \midrule
        \textbf{Q:} Take the last letters of the words in "Larry Page" and concatenate them. \\
        \textbf{A:} Evidence triples: \\
        1. (Larry, last latter, y) \\
        2. (Page, last latter, e) \\
        3. (final answer, is, ye) \\
        Explanation hints: The last letter of "Larry" is "y". The last letter of "Page" is "e". Concatenating them is "ye". \\
        So the answer is ye. \\
        \midrule
        \textbf{Q:} Take the last letters of the words in "Sergey Brin" and concatenate them. \\
        \textbf{A:} Evidence triples: \\
        1. (Sergey, last latter, y) \\
        2. (Brin, last latter, n) \\
        3. (final answer, is, yn) \\
        Explanation hints: The last letter of "Sergey" is "y". The last letter of "Brin" is "n". Concatenating them is "yn".  \\
        So the answer is yn. \\
        \midrule
        \textbf{Q:} Take the last letters of the words in "Bill Gates" and concatenate them. \\
        \textbf{A:} Evidence triples: \\
        1. (Bill, last latter, l) \\
        2. (Gates, last latter, s) \\
        3. (final answer, is, ls) \\
        Explanation hints: The last letter of "Bill" is "l". The last letter of "Gates" is "s". Concatenating them is "ls".   \\
        So the answer is ls. \\
        \bottomrule
    \end{tabular}
    \caption{
    Examples Chain-of-Knowledge prompting for the letter concatenation task.
    }
    \label{tab:appendix-letter-prompt}
\end{table}
\endgroup

% \begingroup
% \begin{table}[H]
%     \centering
%     \small
%     \caption{
%     Few-shot exemplars for full chain of thought prompt for the last letter concatenation task.
%     }
%     \begin{tabular}{p{\linewidth}}
%         \toprule
%         \underline{\textbf{\textsc{Prompt for Last Letter Concatenation}}} \\
%         \textbf{Q:} Take the last letters of the words in "Elon Musk" and concatenate them. \\
%         \textbf{A:} \hl{The last letter of "Elon" is "n". The last letter of "Musk" is "k". Concatenating them is "nk".} The answer is nk. \\
%         \textbf{Q:} Take the last letters of the words in "Larry Page" and concatenate them. \\
%         \textbf{A:} \hl{The last letter of "Larry" is "y". The last letter of "Page" is "e". Concatenating them is "ye".} The answer is ye. \\
%         \textbf{Q:} Take the last letters of the words in "Sergey Brin" and concatenate them. \\
%         \textbf{A:} \hl{The last letter of "Sergey" is "y". The last letter of "Brin" is "n". Concatenating them is "yn".}  The answer is yn. \\
%         \textbf{Q:} Take the last letters of the words in "Bill Gates" and concatenate them. \\
%         \textbf{A:} \hl{The last letter of "Bill" is "l". The last letter of "Gates" is "s". Concatenating them is "ls".} The answer is ls. \\
%         \bottomrule
%     \end{tabular}
%     \label{tab:appendix-letter-concat-prompt}
% \end{table}
% \endgroup

\begingroup
\begin{table}[H]
    \centering
    \small
    \begin{tabular}{p{\linewidth}}
        \toprule
        % \underline{\textbf{\textsc{Coin Flip}}} \\
        \textbf{Q:} A coin is heads up. Ka flips the coin. Sherrie flips the coin. Is the coin still heads up? \\
        \textbf{A:} Evidence triples: \\
        1. (coin, start with, head up) \\
        2. (coin, flips, flipped) \\
        3. (coin, not flips, flipped) \\
        Explanation hints:The coin was flipped by Maybelle. So the coin was flipped 1 time, which is an odd number. The coin started heads up, so after an odd number of flips, it will be tails up. \\
        So the answer is yes. \\
        \midrule
        \textbf{Q:} A coin is heads up. Jamey flips the coin. Teressa flips the coin. Is the coin still heads up? \\
        \textbf{A:} Evidence triples: \\
        1. (coin, start, head up) \\
        2. (coin, flips, flipped) \\
        3. (coin, flips, head up) \\
        Explanation hints: The coin was flipped by Jamey and Teressa. So the coin was flipped 2 times, which is an even number. The coin started heads up, so after an even number of flips, it will still be heads up.  \\
        So the answer is yes. \\
        \midrule
        \textbf{Q:} A coin is heads up. Maybelle flips the coin. Shalonda does not flip the coin. Is the coin still heads up? \\
        \textbf{A:} Evidence triples: \\
        1. (coin, start with, head up)
        2. (coin, flips, flipped)
        3. (coin, not flips, flipped)
        Explanation hints: The coin was flipped by Maybelle. So the coin was flipped 1 time, which is an odd number. The coin started heads up, so after an odd number of flips, it will be tails up.  \\
        So the answer is no. \\
        \midrule
        \textbf{Q:} A coin is heads up. Millicent does not flip the coin. Conception flips the coin. Is the coin still heads up? \\
        \textbf{A:} Evidence triples: \\
        1. (coin, start with, head up) \\
        2. (coin, flips, flipped) \\
        3. (coin, not flips, flipped) \\
        Explanation hints: The coin was flipped by Conception. So the coin was flipped 1 time, which is an odd number. The coin started heads up, so after an odd number of flips, it will be tails up.   \\
        So the answer is no. \\
        \midrule
        \emph{(The rest demonstration examples are shown in Table~\ref{tab:appendix-coin-prompt1_2}.)} \\
        \bottomrule
    \end{tabular}
    \caption{
    Examples Chain-of-Knowledge prompting for the coinflip task.
    }
    \label{tab:appendix-coin-prompt1}
\end{table}
\endgroup

\begingroup
\begin{table}[H]
    \centering
    \small
    \begin{tabular}{p{\linewidth}}
        \toprule
        \emph{(The front demonstration examples are shown in Table~\ref{tab:appendix-coin-prompt1}.)} \\
        \midrule
        \textbf{Q:} A coin is heads up. Sal flips the coin. Raymond does not flip the coin. Is the coin still heads up? \\
        \textbf{A:} Evidence triples: \\
        1. (coin, start with, head up) \\
        2. (coin, flips, flipped) \\
        3. (coin, not flips, flipped) \\
        Explanation hints: The coin was flipped by Sal. So the coin was flipped 1 time, which is an odd number. The coin started heads up, so after an odd number of flips, it will be tails up.    \\
        So the answer is no. \\
        \midrule
        \textbf{Q:} A coin is heads up. Conception flips the coin. Kristian does not flip the coin. Is the coin still heads up? \\
        \textbf{A:} Evidence triples: \\
        1. (coin, start with, head up) \\
        2. (coin, flips, flipped) \\
        3. (coin, not flips, flipped) \\
        Explanation hints: The coin was flipped by Conception. So the coin was flipped 1 time, which is an odd number. The coin started heads up, so after an odd number of flips, it will be tails up. \\
        So the answer is no. \\
        \midrule
        \textbf{Q:} A coin is heads up. Inga does not flip the coin. Elanor does not flip the coin. Is the coin still heads up? \\
        \textbf{A:} Evidence triples: \\
        1. (coin, start with, head up) \\
        2. (coin, not flips, head up) \\
        3. (coin, not flips, head up) \\
        Explanation hints: The coin was flipped by no one. So the coin was flipped 0 times. The coin started heads up, and it was not flipped, so it is still heads up.  \\
        So the answer is yes. \\
        \bottomrule
    \end{tabular}
    \caption{
    (Following Table~\ref{tab:appendix-coin-prompt1}) Examples Chain-of-Knowledge prompting for the coinflip task.
    }
    \label{tab:appendix-coin-prompt1_2}
\end{table}
\endgroup

\begingroup
\begin{table}[H]
    \centering
    \small
    \begin{tabular}{p{\linewidth}}
        \toprule
        \textbf{Q:} Is the following sentence plausible? 'Fernando Tatis Jr. walked on ball four.' \\
        \textbf{A:} Evidence triples: \\
        1. (Fernando Tatis Jr., isA, baseball player) \\
        2. (walking on ball four, is part of, baseball) \\
        Explanation hints: A walk is when the batter reaches first base without hitting the ball. Reaching first base can happen a few ways, but one of them is by the pitcher throwing four balls. \\
        So the answer is yes. \\
        \midrule
        \textbf{Q:} Is the following sentence plausible? 'Jayson Tatum was caught offsides in the Champions Leage Semifinal.' \\
        \textbf{A:} Evidence triples: \\
        1. (Jayson Tatum, isA, basketball player) \\
        2. (being caught offsides, is part of, soccer) \\
        Explanation hints: Being caught offsides is part of soccer, yet, Jayson Tatum is a basketball player.  \\
        So the answer is no. \\
        \midrule
        \textbf{Q:} Is the following sentence plausible? 'Marouane Fellaini scored in extra time.' \\
        \textbf{A:} Evidence triples: \\
        1. (Marouane Fellaini, isA, soccer player) \\
        2. (scoring in extra time, is part of, soccer) \\
        So, the answer is yes. \\
        \midrule
        \textbf{Q:} Is the following sentence plausible? 'Gerrit Cole committed a handball in the European Cup.' \\
        \textbf{A:} Evidence triples: \\
        1. (Gerrit Cole, isA, soccer player) \\
        2. (committing a handball, is part of, soccer) \\
        Explanation hints: Committing a handball is part of soccer, yet, Gerrit Cole is a soccer player.  \\
        So the answer is no. \\
        \midrule
        \textbf{Q:} Is the following sentence plausible? 'Klaas Jan Huntelaar scored the easy layup.' \\
        \textbf{A:} Evidence triples: \\
        1. (Klaas Jan Huntelaar, isA, soccer player) \\
        2. (scoring a layup, is part of, basketball) \\
        Explanation hints: Scoring a layup is part of basketball, yet, Klaas Jan Huntelaar is a soccer player.   \\
        So the answer is no. \\
        \midrule
        \textbf{Q:} Is the following sentence plausible? 'Mario Gomez earned a direct kick.' \\
        \textbf{A:} Evidence triples: \\
        1. (Mario, isA, soccer player) \\
        2. (earning a direct kick, is part of, soccer) \\
        Explanation hints: Earning a direct kick is part of soccer, and Mario Gomez is a soccer player.    \\
        So the answer is yes. \\
        \bottomrule
    \end{tabular}
    \caption{
    Examples Chain-of-Knowledge prompting for the sports understanding task.
    }
    \label{tab:appendix-sports-prompt}
\end{table}
\endgroup

\begingroup
\begin{table}[H]
    \centering
    \small
    \begin{tabular}{p{\linewidth}}
        \toprule
\textbf{Q:} Putting a cardboard box in a bin instead of the trash can? \\
Answer Choices: (A) conserve energy for later (B) save wild animal species (C) keep it from the trash (D) reduce the height of landfills \\
\textbf{A:} Evidence triples: \\
1. (cardboard box, isA, item) \\
2. (bin, isA, container) \\
3. (cardboard box, usage, container) \\
4. (trash can, usage, container) \\
5. (bin, usage, container) \\
Explanation hints: The answer must be an environmental reason. Of the above choices, only (D) are environmental reasons.\\
So the answer is (D). \\
\midrule
\textbf{Q:} When the needle of a compass lines up with Earth 's magnetic poles , the needle is? \\
Answer Choices: (A) pointing east (B) pointing west (C) pointing at arctic (D) pointing to space \\
\textbf{A:} Evidence triples: \\
1. (compass, is affected by, Earth's magnetic field) \\
2. (magnetic field, go, from the south pole to the north pole) \\
Explanation hints: The needle of the compass is affected by Earth's magnetic field. The magnetic field lines go from the south pole to the north pole. So, when the needle of the compass is lined up with the Earth's magnetic poles, the needle is pointing north. \\
So the answer is (C). \\
\midrule
\textbf{Q:} Which uses gills to breathe? \\
Answer Choices: (A) hermit crab (B) human (C) blue whale (D) bluebird \\
\textbf{A:} Evidence triples: \\
1. (gills, isA, respiratory organ) \\
2. (respiratory organ, function, breathing) \\
3. (hermit crab, hasA, gills) \\
4. (blue whale, hasA, gills) \\
5. (bluebird, doesNotHave, gills) \\
Explanation hints: The answer must be an item that has gills. Gills are respiratory organs that help with breathing. Of the above choices, only hermit crabs and blue whales have gills. \\
So the answer is (A). \\
\midrule
\textbf{Q:} Which is the most accurate statement? \\
Answer Choices: (A) finches with fat beaks can eat ants in a small anthill (B) finches with narrow beaks can eat only human meat (C) finches with narrow beaks can eat ants in a small anthill (D) finches with fat beaks can eat only human meat \\
\textbf{A:} Evidence triples: \\
1. (finches, isA, bird)\\
2. (bird, hasA, beak)\\
3. (finches, hasA, fat beak)\\
4. (finches, hasA, narrow beak)\\
5. (finches, eats, ants)\\
6. (finches, eats, human meat)\\
Explanation hints: The answer must be a statement about what finches can eat. Of the above choices, only finches with narrow beaks can eat ants in a small anthill.\\
So the answer is (C).\\
% \midrule
        \bottomrule
    \end{tabular}
    \caption{
    Examples Chain-of-Knowledge prompting for the ARC-c task.
    }
    \label{tab:appendix-arc-prompt}
\end{table}
\endgroup

\begingroup
\begin{table*}[H]
    \centering
    \small
    \begin{tabular}{p{\linewidth}}
    \toprule
\textbf{Q:} What type of useful product can be made from the moving winds? \\
Answer Choices: (A) metal (B) wood (C) bananas (D) electricity \\
\textbf{A:} Evidence triples: \\ 
1. (wind, isA, moving air) \\ 
2. (electricity, isA, moving electrons)\\ 
3. (wind, can generate, electricity)\\ 
Explanation hints: The answer must be a type of energy. Of the above choices, only electricity can be made from the moving winds.\\ 
So the answer is (D).\\ 
\midrule
\textbf{Q: }Harriet wants to know the area of a rectangular sheet of paper. Its size is unknown to her. Which one of these will be the only tool that she needs? \\ 
Answer Choices: (A) a ruler (B) a compass (C) a calculator (D) a protractor \\ 
\textbf{A: }Evidence triples: \\ 
1. (area, isA, measurement) \\ 
2. (rectangular sheet of paper, isA, two-dimensional object)\\ 
3. (area, measurement of, two-dimensional object)\\ 
4. (ruler, usage, measuring)\\ 
5. (ruler, usage, measuring length)\\ 
6. (ruler, usage, measuring width)\\ 
7. (area, calculation of, length x width)\\ 
Explanation hints: The answer must be an item that can help Harriet find the area of the rectangular sheet of paper. Of the above choices, only a ruler can help Harriet find the area of the rectangular sheet of paper.\\ 
So the answer is (A).\\ 
\midrule
\textbf{Q:} What uses eyes to see by sensing light? \\
Answer Choices: (A) trees (B) algae (C) worms (D) elephants \\
\textbf{A:} Evidence triples: \\
1. (eyes, isA, organ) \\
2. (eyes, usage, seeing) \\
3. (eyes, sensing, light) \\
4. (trees, usage, photosynthesis) \\
5. (algae, usage, photosynthesis) \\
6. (worms, usage, burrowing) \\
7. (elephants, usage, trampling) \\
Explanation hints: The answer must be an entity that uses eyes to see by sensing light. Of the above choices, only trees, algae, and worms use eyes to see by sensing light.
So the answer is (D). \\
\midrule
\textbf{Q:} The cell of a plant when photosynthesizing may most require? \\
Answer Choices: (A) access to roofless space (B) a complete vacuum in space (C) a room with blackened windows (D) dark basement living quarters \\
\textbf{A:} Evidence triples: \\
1. (cell, isA, plant) \\
2. (photosynthesis, isA, process) \\
3. (cell, requires, photosynthesis) \\
4. (photosynthesis, requires, light) \\
5. (dark basement living quarters, blocks, light) \\
Explanation hints: The cell of a plant when photosynthesizing may most require access to sunlight. Of the above choices, only (A) provides access to sunlight. \\
So the answer is (A). \\
        \bottomrule
    \end{tabular}
     \caption{
    Examples Chain-of-Knowledge prompting for the ARC-c task.
    }
    \label{tab:appendix-arc-prompt2}
\end{table*}
\endgroup

% \begingroup
\begin{table}[H]
    \centering
    \small
    \resizebox{\linewidth}{!}{
    \begin{tabular}{p{\linewidth}}
    \midrule
\textbf{Q:} There are 15 trees in the grove. Grove workers will plant trees in the grove today. After they are done, there will be 21 trees. How many trees did the grove workers plant today? \\
Answer Choices: (A) 12 (B) 6 (C) 8 (D) 14 \\
\textbf{A:} Evidence triples: \\
1. (grove, originally have, 15 trees) \\
2. (grove, now have, 21 trees) \\
3. (21 - 15, equal, 6) \\
Explanation hints: There are 15 trees originally. Then there were 21 trees after some more were planted. So there must have been 21 - 15 = 6.  \\
So the answer is (B). \\
\midrule
\textbf{Q:} If there are 3 cars in the parking lot and 2 more cars arrive, how many cars are in the parking lot? \\
Answer Choices: (A) 3 (B) 2 (C) 5 (D) 1 \\
A: Evidence triples: \\
1. (parking lot, originally have, 3 cars) \\
2. (parking lot, add, 2 cars) \\
3. (parking lot, now have, 3 + 2 = 5 cars) \\
Explanation hints: There are originally 3 cars. 2 more cars arrive. 3 + 2 = 5. \\
So the answer is (C). \\
\midrule
\textbf{Q:} Leah had 32 chocolates and her sister had 42. If they ate 35, how many pieces do they have left in total? \\
Answer Choices: (A) 39 (B) 74 (C) 109 (D) 40 \\
\textbf{A:} Evidence triples: \\
1. (Leah, have, 32 chocolates) \\
2. (her sister, have, 42 chocolates) \\
3. (they, have, 32 + 42 = 74 chocolates) \\
4. (they, eat, 35 chocolates) \\
5. (thay, now have, 74 - 35 = 39 chocolates) \\
Explanation hints: Originally, Leah had 32 chocolates. Her sister had 42. So in total they had 32 + 42 = 74. After eating 35, they had 74 - 35 = 39.  \\
So the answer is (A).\\
\midrule
\textbf{Q:} Jason had 20 lollipops. He gave Denny some lollipops. Now Jason has 12 lollipops. How many lollipops did Jason give to Denny? \\
Answer Choices: (A) 32 (B) 16 (C) 9 (D) 8 \\
\textbf{A:} Evidence triples: \\
1. (Jason, originally have, 20 lollipops) \\
2. (Jason, now have, 12 lollipops) \\
3. (Jason, give, 20 - 12 = 8 lollipops) \\
Explanation hints: Jason started with 20 lollipops. Then he had 12 after giving some to Denny. So he gave Denny 20 - 12 = 8.  \\
So the answer is (D). \\
        \bottomrule
    \end{tabular}
    }
    \caption{
    Examples Chain-of-Knowledge prompting for the AQuA, GSM8K and MultiArith task.
    }
    \label{tab:appendix-letter-aqua}
\end{table}
% \endgroup

% \begingroup
\begin{table}[H]
    \centering
    \small
    \begin{tabular}{p{\linewidth}}
    \toprule
\textbf{Q:} Shawn has five toys. For Christmas, he got two toys each from his mom and dad. How many toys does he have now? \\
Answer Choices: (A) 5 (B) 20 (C) 9 (D) 1 \\
\textbf{A:} Evidence triples: \\
1. (Shawn, have, 5 toys) \\
2. (his mon, give him, 2 toys) \\
3. (his dad, give him, 2 toys) \\
4. (Shawn, now have, 5 + 2 + 2 = 9 toys) \\
Explanation hints: Shawn started with 5 toys. If he got 2 toys each from his mom and dad, then that is 4 more toys. 5 + 4 = 9. \\
So the answer is (C). \\
\midrule
\textbf{Q:} There were nine computers in the server room. Five more computers were installed each day, from monday to thursday. How many computers are now in the server room? \\
Answer Choices: (A) 20 (B) 29 (C) 11 (D) 18 \\
\textbf{A:} Evidence triples: \\
1. (server room, originally have, 9 computers) \\
2. (each day, installed, 5 computers) \\
3. (each from monday to thursday, have, 4 days) \\
4. (5 * 4, equal, 20) \\
5. (server room, now have, 9 + 20 = 29) \\
Explanation hints: There were originally 9 computers. For each of 4 days, 5 more computers were added. So 5 * 4 = 20 computers were added. 9 + 20 is 29. \\
So the answer is (B). \\
\midrule
\textbf{Q:} Michael had 58 golf balls. On tuesday, he lost 23 golf balls. On wednesday, he lost 2 more. How many golf balls did he have at the end of wednesday? \\
Answer Choices: (A) 33 (B) 35 (C) 81 (D) 83 \\
\textbf{A:} Evidence triples: \\
1. (Michael, have, 58 golf balls) \\
2. (Michael, lost, 23 golf balls) \\
3. (Michael, lost, 2 golf balls) \\
4. (Michael, now have, 58 - 23 - 2 = 33 golf balls) \\
Explanation hints: Michael started with 58 golf balls. After losing 23 on tuesday, he had 58 - 23 = 35. After losing 2 more, he had 35 - 2 = 33 golf balls.  \\
So the answer is (A). \\
\midrule
\textbf{Q:} Olivia has \$23. She bought five bagels for \$3 each. How much money does she have left? \\
Answer Choices: (A) 37 (B) 8 (C) 15 (D) 10 \\
\textbf{A:} Evidence triples: \\
1. (Olivia, have, 23 dollars) \\
2. (Olivia, buy, 5 * 3 = 15 dollars) \\
3. (Olivia, now have, 23 - 15 = 8 dollars) \\
Explanation hints: Olivia had 23 dollars. 5 bagels for 3 dollars each will be 5 x 3 = 15 dollars. So she has 23 - 15 dollars left. 23 - 15 is 8.  \\
So the answer is (B). \\

        \bottomrule
    \end{tabular}
    \caption{
    Examples Chain-of-Knowledge prompting for the AQuA, GSM8K and MultiArith task.
    }
    \label{tab:appendix-letter-aqua}
\end{table}
% \endgroup

\begingroup
\begin{table}[H]
    \centering
    \small
    \begin{tabular}{p{\linewidth}}
\toprule
\textbf{Q:} Do hamsters provide food for any animals? \\
\textbf{A:} Evidence triples: \\
1. (Hamsters, isA, prey animals) \\
2. (Prey, is food for, predators) \\
3. (hamsters, provide food, animals) \\
Explanation hints: Hamsters are prey animals. Prey are food for predators. Thus, hamsters provide food for some animals.  \\
So the answer is yes. \\
\midrule
\textbf{Q:} Could Brooke Shields succeed at University of Pennsylvania? \\
\textbf{A:} Evidence triples: \\
1. (Brooke Shields, isA, student) \\
2. (student, could succeed, at University of Pennsylvania) \\
3. (Brooke Shields, could succeed, at University of Pennsylvania) \\
Explanation hints: Brooke Shields is a student. Students could succeed at University of Pennsylvania. Thus, Brooke Shields could succeed at University of Pennsylvania. \\
So the answer is yes. \\
\midrule
\textbf{Q:} Yes or no: Hydrogen’s atomic number squared exceeds number of Spice Girls? \\
\textbf{A:} Evidence triples: \\
1. (Hydrogen, has atomic number, 1) \\
2. (1, squared is, 1) \\
3. (1, exceeds, 5) \\
4. (5, is the number of, Spice Girls) \\
Explanation hints: Hydrogen has an atomic number of 1. 1 squared is 1. There are 5 Spice Girls. Thus, Hydrogen’s atomic number squared is less than 5.  \\
So the answer is no. \\
\midrule
\textbf{Q:} Yes or no: Is it common to see frost during some college commencements? \\
\textbf{A:} Evidence triples: \\
1. (Frost, isA, weather condition) \\
2. (Weather condition, is, common during some college commencements) \\
3. (Frost, is common, during some college commencements) \\
Explanation hints: Frost is a weather condition. Weather conditions are common during some college commencements. Thus, frost is common during some college commencements. \\
So the answer is yes. \\
\midrule
\textbf{Q:} Yes or no: Could a llama birth twice during War in Vietnam (1945-46)? \\
\textbf{A:} Evidence triples: \\
1. (Llama, isA, mammal) \\
2. (Mammal, gives birth, live young) \\
3. (Llama, could give birth, during War in Vietnam) \\
Explanation hints: Llamas are mammals. Mammals give birth to live young. Therefore, it is possible for a llama to give birth during the War in Vietnam. \\
So the answer is no. \\
\midrule
\textbf{Q:} Yes or no: Would a pear sink in water? \\
\textbf{A:} Evidence triples: \\
1. (pear, density, 0.6g/cm3) \\
2. (water, density, 1.0g/cm3) \\
3. (1.0g/cm3, is larger than, 0.6g/cm3) \\
4. (pear, can not sink in, water) \\
Explanation hints: The density of a pear is about 0.6g/cm3, which is less than water. Objects less dense than water float. Thus, a pear would float. \\
So the answer is no. \\
        \bottomrule
    \end{tabular}
    \caption{
    Examples Chain-of-Knowledge prompting for the BoolQ task.
    }
    \label{tab:appendix-letter-boolq}
\end{table}
\endgroup

\begingroup
\begin{table}[H]
    \centering
    \small
    \begin{tabular}{p{\linewidth}}
\toprule
\textbf{Q:} What do people use to absorb extra ink from a fountain pen? \\
Answer Choices: (A) shirt pocket (B) calligrapher’s hand (C) inkwell (D) desk drawer (E) blotter \\
\textbf{A:} Evidence triples: \\
1. (fountain pen, isA, item) \\
2. (ink, isA, liquid) \\
3. (fountain pen, carrier of, liquid) \\
4. (fountain pen, usage, writing) \\
5. (blotters, usage, writing) \\
6. (blotters, absorb, liquid) \\
Explanation hints: The answer must be an item that can absorb ink. A fountain pen which is full of liquid can writing on a blotter. Of the above choices, only blotters are used to absorb ink. \\
So the answer is (E). \\
\midrule
\textbf{Q:} What home entertainment equipment requires cable? \\
Answer Choices: (A) radio shack (B) substation (C) television (D) cabinet \\
\textbf{A:} Evidence triples: \\
1. (home entertainment equipment, isA, item) \\
2. (cable, isA, item) \\ 
3. (home entertainment equipment, requires, cable) \\
4. (television, isA, home entertainment equipment) \\
5. (television, requires, cable) \\
Explanation hints: The answer must be an item of home entertainment equipment that requires cable. Of the above choices, only television requires cable. \\
So the answer is (C). \\
\midrule
\textbf{Q:} The fox walked from the city into the forest, what was it looking for? \\
Answer Choices: (A) pretty flowers (B) hen house (C) natural habitat (D) storybook \\
\textbf{A:} Evidence triples: \\
1. (fox, isA, animal) \\
2. (forest, place of residence, animal) \\
3. (natural habitat, located in, forest) \\
Explanation hints: The answer must be something in the forest. Of the above choices, only natural habitat is in the forest which is the living place for fox animal. \\
So the answer is (C). \\
        \midrule
        \emph{(The rest demonstration examples are shown in Table~\ref{tab:appendix-letter-csqa1_2}.)} \\
        \bottomrule
    \end{tabular}
    \caption{
    Examples of Chain-of-Knowledge prompting for the CSQA task.
    }
    \label{tab:appendix-letter-csqa1}
\end{table}
\endgroup

\begingroup
\begin{table}[H]
    \centering
    \small
    \begin{tabular}{p{\linewidth}}
\toprule
\emph{(The front demonstration examples are shown in Table~\ref{tab:appendix-letter-csqa1}.)} \\
\midrule
\textbf{Q:} Sammy wanted to go to where the people were. Where might he go? \\
Answer Choices: (A) populated areas (B) race track (C) desert (D) apartment (E) roadblock \\
\textbf{A: }Evidence triples: \\
1. (Sammy, isA, person) \\
2. (populated areas, place of residence, people) \\
Explanation hints: Sammy is a person, so that the answer must be a place with a lot of people. Of the above choices, only populated areas have a lot of people. \\
So the answer is (A). \\
\midrule
\textbf{Q:} Where do you put your grapes just before checking out? \\
Answer Choices: (A) mouth (B) grocery cart (C) super market (D) fruit basket (E) fruit market \\
\textbf{A: }Evidence triples: \\
1. (checking out, isA, action) \\
2. (checking out, place of take place, mall) \\
3. (grapes, isA, merchandise) \\
4. (mall, sell, fruit) \\
5. (grocery cart, usage, hold merchandise before checking out) \\
Explanation hints: The answer should be the place where grocery items are placed before checking out. Of the above choices, grocery cart makes the most sense for holding grocery items. \\
So the answer is (B). \\
\midrule
\textbf{Q:} Google Maps and other highway and street GPS services have replaced what? \\
Answer Choices: (A) united states (B) mexico (C) countryside (D) atlas \\
\textbf{A:} Evidence triples: \\
1. (Google Maps, isA, webapp) \\
2. (GPS, isA, navigation systems) \\
3. (atlas, usage, navigation) \\
Explanation hints: The answer must be something that used to do what Google Maps and GPS services do, which is to give directions. Of the above choices, only atlases are used to give directions. \\
So the answer is (D). \\
        \bottomrule
    \end{tabular}
    \caption{
    (Following Table~\ref{tab:appendix-letter-csqa1}) Examples of Chain-of-Knowledge prompting for the CSQA task.
    }
    \label{tab:appendix-letter-csqa1_2}
\end{table}
\endgroup

\begingroup
\begin{table}[H]
    \centering
    \small
    \begin{tabular}{p{\linewidth}}
    \toprule
\textbf{Q:} There are 15 trees in the grove. Grove workers will plant trees in the grove today. After they are done, there will be 21 trees. How many trees did the grove workers plant today? \\
\textbf{A:} Explanation hints: There are 15 trees originally. Then there were 21 trees after some more were planted. So there must have been 21 - 15 = 6.  \\
Evidence triples: \\
1. (grove, originally have, 15 trees) \\
2. (grove, now have, 21 trees) \\
3. (21 - 15, equal, 6) \\
So the answer is 6. \\
\midrule
\textbf{Q:} If there are 3 cars in the parking lot and 2 more cars arrive, how many cars are in the parking lot?  \\
\textbf{A:} Explanation hints: There are originally 3 cars. 2 more cars arrive. 3 + 2 = 5. \\
Evidence triples: \\
1. (parking lot, originally have, 3 cars) \\
2. (parking lot, add, 2 cars) \\
3. (parking lot, now have, 3 + 2 = 5 cars) \\
So the answer is 5. \\
\midrule
\textbf{Q:} Leah had 32 chocolates and her sister had 42. If they ate 35, how many pieces do they have left in total? \\
\textbf{A:} Explanation hints: Originally, Leah had 32 chocolates. Her sister had 42. So in total they had 32 + 42 = 74. After eating 35, they had 74 - 35 = 39. \\
Evidence triples: \\
1. (Leah, have, 32 chocolates) \\
2. (her sister, have, 42 chocolates) \\
3. (they, have, 32 + 42 = 74 chocolates) \\
4. (they, eat, 35 chocolates) \\
5. (thay, now have, 74 - 35 = 39 chocolates) \\
So the answer is 39. \\
\midrule
\emph{(The rest demonstration examples are shown in Table~\ref{tab:appendix-letter-svamp_2}.)} \\
    \bottomrule
    \end{tabular}
    \caption{
    Examples Chain-of-Knowledge prompting for the SVAMP task.
    }
    \label{tab:appendix-letter-svamp}
\end{table}
\endgroup

\begingroup
\begin{table}[H]
    \centering
    \small
    \begin{tabular}{p{\linewidth}}
    \toprule
    \emph{(The front demonstration examples are shown in Table~\ref{tab:appendix-letter-svamp}.)} \\
    \midrule
\textbf{Q:} Jason had 20 lollipops. He gave Denny some lollipops. Now Jason has 12 lollipops. How many lollipops did Jason give to Denny? \\
\textbf{A:} Explanation hints: Jason started with 20 lollipops. Then he had 12 after giving some to Denny. So he gave Denny 20 - 12 = 8.  v
Evidence triples: \\
1. (Jason, originally have, 20 lollipops) \\
2. (Jason, now have, 12 lollipops) \\ 
3. (Jason, give, 20 - 12 = 8 lollipops) \\
So the answer is 8. \\
\midrule
\textbf{Q:} Shawn has five toys. For Christmas, he got two toys each from his mom and dad. How many toys does he have now? \\
\textbf{A:} Explanation hints: Shawn started with 5 toys. If he got 2 toys each from his mom and dad, then that is 4 more toys. 5 + 4 = 9.  \\
Evidence triples: \\
1. (Shawn, have, 5 toys) \\
2. (his mon, give him, 2 toys) \\
3. (his dad, give him, 2 toys) \\
4. (Shawn, now have, 5 + 2 + 2 = 9 toys) \\
So the answer is 9. \\
\midrule
\textbf{Q:} There were nine computers in the server room. Five more computers were installed each day, from monday to thursday. How many computers are now in the server room? \\
\textbf{A:} Explanation hints: There were originally 9 computers. For each of 4 days, 5 more computers were added. So 5 * 4 = 20 computers were added. 9 + 20 is 29.  \\
Evidence triples: \\
1. (server room, originally have, 9 computers) \\
2. (each day, installed, 5 computers) \\
3. (each from monday to thursday, have, 4 days) \\
4. (5 * 4, equal, 20) \\
5. (server room, now have, 9 + 20 = 29) \\
So the answer is 29. \\
    \bottomrule
    \end{tabular}
    \caption{
    (Follow by Table~\ref{tab:appendix-letter-svamp}) Examples Chain-of-Knowledge prompting for the SVAMP task.
    }
    \label{tab:appendix-letter-svamp_2}
\end{table}
\endgroup

\begingroup
\begin{table}[H]
    \centering
    \small
    \begin{tabular}{p{\linewidth}}
    \toprule
\textbf{Q: }Putting a cardboard box in a bin instead of the trash can? \\
Answer Choices: (A) conserve energy for later (B) save wild animal species (C) keep it from the trash (D) reduce the height of landfills \\
\textbf{A: }Evidence triples: \\
1. (cardboard box, isA, item) \\
2. (bin, isA, container) \\
3. (cardboard box, usage, container) \\
4. (trash can, usage, container) \\
5. (bin, usage, container) \\
Explanation hints: The answer must be an environmental reason. Of the above choices, only (D) are environmental reasons. \\
So the answer is (D). \\
\midrule
\textbf{Q:} When the needle of a compass lines up with Earth 's magnetic poles , the needle is? \\
Answer Choices: (A) pointing east (B) pointing west (C) pointing at arctic (D) pointing to space \\
\textbf{A:} Evidence triples: \\
1. (compass, is affected by, Earth's magnetic field) \\
2. (magnetic field, go, from the south pole to the north pole) \\
Explanation hints: The needle of the compass is affected by Earth's magnetic field. The magnetic field lines go from the south pole to the north pole. So, when the needle of the compass is lined up with the Earth's magnetic poles, the needle is pointing north.  \\
So the answer is (C). \\
\midrule
\textbf{Q: }Which uses gills to breathe? \\
Answer Choices: (A) hermit crab (B) human (C) blue whale (D) bluebird \\
\textbf{A: }Evidence triples: \\
1. (gills, isA, respiratory organ) \\
2. (respiratory organ, function, breathing) \\
3. (hermit crab, hasA, gills) \\
4. (blue whale, hasA, gills) \\
5. (bluebird, doesNotHave, gills) \\
Explanation hints: The answer must be an item that has gills. Gills are respiratory organs that help with breathing. Of the above choices, only hermit crabs and blue whales have gills. \\
So the answer is (A). \\
\midrule
\textbf{Q:} Which is the most accurate statement? \\
Answer Choices: (A) finches with fat beaks can eat ants in a small anthill (B) finches with narrow beaks can eat only human meat (C) finches with narrow beaks can eat ants in a small anthill (D) finches with fat beaks can eat only human meat \\
A: Evidence triples: \\
1. (finches, isA, bird) \\ 
2. (bird, hasA, beak) \\
3. (finches, hasA, fat beak) \\
4. (finches, hasA, narrow beak) \\
5. (finches, eats, ants) \\
6. (finches, eats, human meat) \\
Explanation hints: The answer must be a statement about what finches can eat. Of the above choices, only finches with narrow beaks can eat ants in a small anthill. \\
So the answer is (C). \\
\midrule
\emph{(The rest demonstration examples are shown in Table~\ref{tab:appendix-letter-OpenbookQA2}).} \\
    \bottomrule
    \end{tabular}
    \caption{
    Examples of Chain-of-Knowledge prompting for the OpenbookQA task.
    }
    \label{tab:appendix-letter-OpenbookQA}
\end{table}
\endgroup

\begingroup
\begin{table}[H]
    \centering
    \small
    \begin{tabular}{p{\linewidth}}
    \toprule
    \emph{(Other demonstration examples are shown in Table~\ref{tab:appendix-letter-OpenbookQA}.)} \\
    \midrule
    \textbf{Q:} What type of useful product can be made from the moving winds? \\
Answer Choices: (A) metal (B) wood (C) bananas (D) electricity \\
\textbf{A: }Evidence triples: \\
1. (wind, isA, moving air) \\
2. (electricity, isA, moving electrons) \\
3. (wind, can generate, electricity) \\
Explanation hints: The answer must be a type of energy. Of the above choices, only electricity can be made from the moving winds. \\
So the answer is (D). \\
\midrule
\textbf{Q:} Harriet wants to know the area of a rectangular sheet of paper. Its size is unknown to her. Which one of these will be the only tool that she needs? \\
Answer Choices: (A) a ruler (B) a compass (C) a calculator (D) a protractor \\
\textbf{A:} Evidence triples: \\
1. (area, isA, measurement) \\
2. (rectangular sheet of paper, isA, two-dimensional object) \\
3. (area, measurement of, two-dimensional object) \\
4. (ruler, usage, measuring) \\
5. (ruler, usage, measuring length) \\
6. (ruler, usage, measuring width) \\
7. (area, calculation of, length x width) \\
Explanation hints: The answer must be an item that can help Harriet find the area of the rectangular sheet of paper. Of the above choices, only a ruler can help Harriet find the area of the rectangular sheet of paper. \\
So the answer is (A). \\
\midrule
\textbf{Q: }What uses eyes to see by sensing light? \\
Answer Choices: (A) trees (B) algae (C) worms (D) elephants \\
\textbf{A:} Evidence triples: \\
1. (eyes, isA, organ) \\
2. (eyes, usage, seeing) \\
3. (eyes, sensing, light) \\
4. (trees, usage, photosynthesis) \\
5. (algae, usage, photosynthesis) \\
6. (worms, usage, burrowing) \\
7. (elephants, usage, trampling) \\ 
Explanation hints: The answer must be an entity that uses eyes to see by sensing light. Of the above choices, only trees, algae, and worms use eyes to see by sensing light. \\
So the answer is (D). \\
\midrule
\textbf{Q: }The cell of a plant when photosynthesizing may most require? \\
Answer Choices: (A) access to roofless space (B) a complete vacuum in space (C) a room with blackened windows (D) dark basement living quarters \\
\textbf{A:} Evidence triples: \\
1. (cell, isA, plant) \\
2. (photosynthesis, isA, process) \\
3. (cell, requires, photosynthesis) \\
4. (photosynthesis, requires, light) \\
5. (dark basement living quarters, blocks, light) \\
Explanation hints: The cell of a plant when photosynthesizing may most require access to sunlight. Of the above choices, only (A) provides access to sunlight. \\
So the answer is (A). \\
    \bottomrule
    \end{tabular}
    \caption{
    (Follow by Table~\ref{tab:appendix-letter-OpenbookQA}) Examples of Chain-of-Knowledge prompting for the OpenbookQA task.
    }
    \label{tab:appendix-letter-OpenbookQA2}
\end{table}
\endgroup

\begingroup
\begin{table}[H]
    \centering
    \small
    \begin{tabular}{p{\linewidth}}
    \toprule
\textbf{Q:} Do hamsters provide food for any animals? \\
\textbf{A:} Evidence triples: \\
1. (Hamsters, isA, prey animals) \\
2. (Prey, is food for, predators) \\
3. (hamsters, provide food, animals) \\
Explanation hints: Hamsters are prey animals. Prey are food for predators. Thus, hamsters provide food for some animals.  \\
So the answer is yes. \\
\midrule
\textbf{Q:} Could Brooke Shields succeed at University of Pennsylvania? \\
\textbf{A:} Evidence triples: \\
1. (Brooke Shields, isA, student) \\
2. (student, could succeed, at University of Pennsylvania) \\
3. (Brooke Shields, could succeed, at University of Pennsylvania) \\
Explanation hints: Brooke Shields is a student. Students could succeed at University of Pennsylvania. Thus, Brooke Shields could succeed at University of Pennsylvania. \\
So the answer is yes. \\
\midrule
\textbf{Q:} Yes or no: Hydrogen’s atomic number squared exceeds number of Spice Girls? \\
\textbf{A:} Evidence triples: \\
1. (Hydrogen, has atomic number, 1) \\
2. (1, squared is, 1) \\
3. (1, exceeds, 5) \\
4. (5, is the number of, Spice Girls) \\
Explanation hints: Hydrogen has an atomic number of 1. 1 squared is 1. There are 5 Spice Girls. Thus, Hydrogen’s atomic number squared is less than 5.  \\
So the answer is no. \\
\midrule
\textbf{Q:} Yes or no: Is it common to see frost during some college commencements? \\
\textbf{A:} Evidence triples: \\
1. (Frost, isA, weather condition) \\
2. (Weather condition, is, common during some college commencements) \\
3. (Frost, is common, during some college commencements) \\
Explanation hints: Frost is a weather condition. Weather conditions are common during some college commencements. Thus, frost is common during some college commencements. \\
So the answer is yes. \\
\midrule
\textbf{Q:} Yes or no: Could a llama birth twice during War in Vietnam (1945-46)? \\
\textbf{A:} Evidence triples: \\
1. (Llama, isA, mammal) \\
2. (Mammal, gives birth, live young) \\
3. (Llama, could give birth, during War in Vietnam) \\
Explanation hints: Llamas are mammals. Mammals give birth to live young. Therefore, it is possible for a llama to give birth during the War in Vietnam. \\
So the answer is no.  \\
\midrule
\textbf{Q:} Yes or no: Would a pear sink in water? \\
\textbf{A:} Evidence triples: \\
1. (pear, density, 0.6g/cm3) \\
2. (water, density, 1.0g/cm3) \\
3. (1.0g/cm3, is larger than, 0.6g/cm3) \\
4. (pear, can not sink in, water) \\
Explanation hints: The density of a pear is about 0.6g/cm3, which is less than water. Objects less dense than water float. Thus, a pear would float.  \\
So the answer is no. \\

    \bottomrule
    \end{tabular}
    \caption{
    Examples of Chain-of-Knowledge prompting for the StrategyQA task.
    }
    \label{tab:appendix-letter-OpenbookQA-1}
\end{table}
\endgroup

\end{document}